%% file: main.tex
\documentclass[reprint,aps,pre]{revtex4-1}

\usepackage{amsmath,amsfonts,amscd,amssymb,graphicx,subeqnar}
\usepackage[bbgreekl]{mathbbol}

\input{z_vardef}
\usepackage{dcolumn}
\usepackage{overpic}
\usepackage{color}
\usepackage{adjustbox}

\newcommand{\revcolor}{black}
\newcommand{\rev}[1]{\textcolor{\revcolor}{#1}}

\usepackage{float}
\usepackage{algorithm}
\usepackage{algpseudocode}
\algnewcommand\algorithmicinput{\textbf{Input:}}
\algnewcommand\Input{\item[\algorithmicinput]}
\usepackage{parskip}
\DeclareSymbolFontAlphabet{\mathbb}{AMSb}
\DeclareSymbolFontAlphabet{\mathbbl}{bbold}

\usepackage{nomencl} 
\usepackage{etoolbox}
\usepackage{comment}
\makenomenclature

\newcommand{\nomunit}[1]{%
\renewcommand{\nomentryend}{\hspace*{\fill}#1}}

\renewcommand{\nomgroup}[1]{%
  \ifstrequal{#1}{L}{\item[\textbf{Latin Symbols}]}{%
  \ifstrequal{#1}{G}{\item[\textbf{Greek Letters}]}{%
  \ifstrequal{#1}{A}{\item[\textbf{Acronyms}]}{}}}}

\renewcommand{\vec}[1]{\mathbf{#1}}

\newcommand{\norma}[1]{\left|\left|{#1}\right|\right|}

\usepackage{hyperref}

\makeatletter
\@ifundefined{sf@counterlist}{\def\sf@counterlist{}}{}
\makeatother

\usepackage[dvipsnames, table]{xcolor}

\definecolor{bluePolimi}{RGB}{22, 44, 80}
\definecolor{lightBluePolimi}{RGB}{91, 122, 172}
\definecolor{redPolimi}{RGB}{180, 0, 0}
\definecolor{greenPolimi}{RGB}{78, 172, 91}
\definecolor{green2}{RGB}{0, 110, 0}

\makeatletter
\hypersetup{colorlinks=true}
\AtBeginDocument{\@ifpackageloaded{hyperref}
  {\def\@linkcolor{blue}
   \def\@anchorcolor{red}
   \def\@citecolor{red}
   \def\@filecolor{red}
   \def\@urlcolor{redPolimi}
   \def\@menucolor{red}
   \def\@pagecolor{cyan}
\begingroup
  \@makeother\`%
  \@makeother\=%
  \edef\x{%
    \edef\noexpand\x{%
      \endgroup
      \noexpand\toks@{%
        \catcode 96=\noexpand\the\catcode`\noexpand\`\relax
        \catcode 61=\noexpand\the\catcode`\noexpand\=\relax
      }%
    }%
    \noexpand\x
  }%
\x
\@makeother\`
\@makeother\=
}{}}
\makeatother

\usepackage{comment}
\usepackage{tabularx}
\usepackage{booktabs} 

\begin{document}

\title{Towards Efficient Parametric State Estimation in Circulating Fuel Reactors with Shallow Recurrent Decoder Networks}
\author{Stefano Riva$^{a,b}$, Carolina Introini$^{b}$, J. Nathan Kutz$^{a,c}$, Antonio Cammi$^{d,b,*}$}
\affiliation{$^{a}$Autodesk Research, 6 Agar Street, London, United Kingdom}
\affiliation{$^{b}$Politecnico di Milano, Department of Energy, CeSNEF - Nuclear Engineering Division, 20156 Milan, Italy}
\affiliation{$^{c}$Department of Applied Mathematics and  Electrical and Computer Engineering, University of Washington, Seattle, WA 98195}
\affiliation{$^{d}$Emirates Nuclear Technology Center (ENTC), Department of Mechanical and Nuclear Engineering, Khalifa University, Abu Dhabi, 127788, United Arab Emirates}
\email{antonio.cammi@polimi.it}

\begin{abstract} 
The recent developments in data-driven methods have paved the way for new methodologies to provide accurate and robust full state reconstruction of engineering systems under any operating conditions. Such approaches can be attractive for next-generation nuclear reactors, which represent particularly challenging applications for sensing and monitoring due to the complexity of the strongly coupled physics involved and the extremely harsh environments. Data-driven techniques can combine different sources of information, including computational proxy models and noisy measurements on the system, to robustly estimate the state. This work leverages the novel Shallow Recurrent Decoder (SHRED) architecture to infer the entire state vector (both neutronics and fluid dynamics fields) of a circulating fuel reactor, the Molten Salt Fast Reactor. The standard architecture is extended to treat parametric time-series data, ensuring the possibility of investigating different scenarios and showing its capabilities to provide an accurate state estimation in various conditions. The proposed methodology adopts possible innovative sensing strategies in new-generation reactors, including out-of-core time-series neutron flux measurements alone, in-core sensing measuring the precursor concentrations or mobile probes advected by the fuel flow. The promising results of this work are further strengthened by the possibility of quantifying the uncertainty associated with the state estimation due to the considerably low training cost of the architecture. The ability of SHRED to deliver real-time, system-wide reconstruction highlights its potential for reactor monitoring, control, and digital twin development in next-generation circulating fuel systems.
\end{abstract}

\maketitle


\section{Introduction}

Mathematical modelling and simulations of fluids have become an invaluable tool for design, optimisation, process monitoring, and control of engineering systems, from chemical to nuclear reactors, as well as for understanding the underlying physical mechanisms that govern complex fluid behaviour \cite{BATTISTINI2021116520}. Fluid flows are often characterised by highly non-linear, multi-scale phenomena, such as turbulence, mixing, and coupled multi-physics interactions: numerical simulation through direct solution of proper Partial Differential Equations (PDEs), obtained from local conservation laws, comes at the expense of very high computational costs, ranging from hours to days and even weeks for more complex systems. This shortcoming limits the direct use of PDEs for multi-query and real-time scenarios \cite{lassila_model_2014, quarteroni2015reduced, rozza_model_2020}, including design and shape optimisation problems or online control and process monitoring. 

To address the trade-off between computational accuracy and cost, innovative modelling techniques falling under the data-driven model reduction umbrella \cite{brunton_data-driven_2022} have been proposed. Firstly, Reduced Order Modelling (ROM) approaches have been studied as a possible solution to lower computational costs while keeping the accuracy of the prediction at a desired level \cite{lassila_model_2014, FICK2018214, rozza_model_2020}. These methods aim to obtain a reduced/latent representation of the PDEs, i.e., the high-dimensional problem or {\em Full Order Model} (FOM), which can be solved in a reasonably low time even on personal computers. One of the most powerful dimensionality reduction methods is the {\em Singular Value Decomposition} (SVD) \cite{rozza_model_2020}, a linear algebra technique which is the foundation for {\em Proper Orthogonal Decomposition} (POD) \cite{sirovich_turbulence_1987-1} and Principal Component Analysis: this technique can extract the dominant spatial features from a series of snapshots, i.e. solutions of the FOM, through the generation of a set of modes, retaining most of the energy/information content of the starting dataset \cite{quarteroni2015reduced}. Application of such techniques in the fluid dynamics world is not a novelty. Indeed, the POD method was used for the first time by Sirovich \cite{sirovich_turbulence_1987-1} to obtain coherent structures for turbulent flows, with the basis functions having physically meaningful interpretability \cite{brunton_data-driven_2022, lassila_model_2014}. Closely related to POD is the {\em Dynamic Mode Decomposition} (DMD)~\cite{schmid2010dynamic, Kutz2016book}, which extends the interpretability of low-rank models by reduction to exponential solutions in time~\cite{ichinaga2024pydmd}, which can be endowed with uncertainty quantification~\cite{sashidhar2022bagging}. Both POD and DMD methods are now extensively used in science and engineering \cite{VOLOSHCHUK2022117620, QIU2022117112}.

With the advancements in Machine Learning (ML) and Artificial Intelligence (AI) methods \cite{brunton_data-driven_2022}, the combination of SVD/POD with ML approaches has become a promising pathway for obtaining a reliable, efficient, and fully data-driven framework for state estimation in engineering systems \cite{gong_efficient_2022, gong_reactor_2024, GKINIS2019371, HU2025121170, KAMEYA2024120097, LI2025122003, MOHAN2024120111}. In particular, the data compression provided by the SVD allows for a much lower training cost of the ML models, thereby requiring less data compared to high-dimensional training \cite{gong_efficient_2022, kutz_shallow_2024}. Furthermore, this paradigm opens new ways to integrate measurements directly collected on the physical system with the background knowledge provided by models \cite{argaud_sensor_2018, riva2024multiphysics, kutz_shallow_2024, Luo2023_PODAE} compared to standard data assimilation algorithms \cite{carrassi_data_2018}, which are plagued by long computational times. To develop feasible physics-based surrogate models for monitoring and control of complex fluid-dynamics engineering systems, the need for dimensionality reduction approaches becomes essential \cite{gong_data-enabled_2022}. 

Among engineering applications, nuclear reactors are typically characterised by complex physics that include turbulent flows and feedback effects between thermal-hydraulics and neutronics (following the evolution of the fission events) \cite{LUZZI20104873}, making the associated mathematical models computationally expensive to solve \cite{DuderstadtHamilton, demaziere_6_2020}. The latter coupling effects play a significant role, both in the design and operation of a nuclear reactor, because the interaction of neutrons with matter is dependent on the temperature; thus, ad-hoc codes have to be developed \cite{aufiero2014development, fiorina_gen-foam_2015, novak2022_cardinal} to model the intrinsic multi-physics nature of nuclear reactors properly. Furthermore, innovative reactor concepts \cite{GenIV-RoadMap} pose additional challenges from the modelling, monitoring, and control point of view compared to traditional designs. For example, Molten Salt Fast Reactors (MSFRs) are designed with a liquid fuel \cite{brovchenko2013optimization, aufiero2014development, CERVI2019379}, making the two-way connection between fluid dynamics and neutronics even stronger: the molten salt will both act as coolant for the reactor, while being internally heated due to the fission event, a feature that may affect the fluid-dynamics behaviour of the fluid itself; this aspect requires a suitable simulation tool, able to tackle the coupling effect between the Navier-Stokes equations for fluid dynamics and the Boltzmann equation for neutronics. Furthermore, the presence of a liquid fuel (and no solid structures within) makes sensor positioning particularly challenging \cite{argaud_sensor_2018, INTROINI2023109538}. The MSFR operates in the fast neutron spectrum, making the in-core environment harsher than that of thermal reactors \cite{ICAPP_plus2023, riva2024robuststateestimationpartial}; additionally, the molten salt operates at high temperature, making the insertion of temperature probes a barely impossible task. All these challenges may be shared by other similar engineering systems, for example, featuring high-temperature fluids subjected to particular heating conditions and for which imaging techniques are not feasible (due to, for instance, the presence of steel tubes). It is worth mentioning that for innovative reactor concepts, full-scale demonstrators are not available, and experimental facilities are often proprietary: therefore, high-fidelity modelling must be taken as reference when exploring the phenomenology of such systems, and multi-query simulations the only feasible way to span various operating and accidental scenarios.

Within this framework and in light of the above challenges, this work discusses the possibility of adopting a combination of SVD and ML to provide an innovative methodology for the accurate and reliable state estimation of the state of a liquid-fuel Generation-IV reactor concept. At first, outcore sensors will be considered; then, a potential innovative concept of a "mobile sensor", that is, tracking a radionuclide advected by the molten salt flow within the core by detecting its emission \cite{ORNL2020}. In particular, this work considers a typical accidental scenario of the MSFR, the Unprotected Loss of Fuel Flow (ULOFF) in which the pump breaks and can no longer provide momentum to the liquid fuel; thus, decay heat must be removed through natural circulation. This work aims to show how machine learning techniques can be efficiently used to generate accurate surrogate models of the systems for control and process monitoring purposes, a necessity for the development of digital twins \cite{Grieves_DT, mohanty_physics-infused_2021, mohanty_development_2021} of nuclear reactors and, more in general, of any complex engineering system.

The selected ML architecture is SHRED, acronym for SHallow REcurrent Decoder \cite{williams2022data, ebers2023leveraging, kutz_shallow_2024}, which is used to map the trajectories of measures of a given observable quantity to the full state space. This technique can be considered as a generalisation of the separation of variables \cite{kutz_shallow_2024, shredrom}, providing more interpretability compared to other deep learning architectures. The SHRED architecture comes with important advantages compared to other ML techniques \cite{williams2022data, kutz_shallow_2024, riva2024robuststateestimationpartial, shredrom}: sensors can be placed randomly and limited to only 3; training occurs in a compressed space obtained with the SVD, thus it can be performed in minutes even on a personal computer avoiding the need for powerful GPUs; most importantly, SHRED requires minimum hyper-parameter tuning, as it has been shown how the same architecture can provide accurate results on a wide range of problems belonging to different fields \cite{kutz_shallow_2024}. Tomasetto et al. \cite{shredrom} formally discussed the architecture for generating parametric reduced order models and compared its performance against other state estimation methods, based on the coupling ML and POD, such as POD-DeepONet \cite{LU2022114778} or POD-Autoencoders \cite{Nair_Goza_2020, Luo2023_PODAE}, showing SHRED to be superior both in terms of accuracy and computational time.

This novel approach has already been applied by the authors in \cite{riva2024robuststateestimationpartial}, focusing on the state reconstruction (neutron fluxes, precursor concentration groups, temperature, pressure and velocity) from outcore sensors for a single-parameter transient scenario. In addition, the authors presented a comparison with the Generalised Empirical Interpolation Method \cite{maday_generalized_2015, introini_stabilization_2023}, a standard well-established state estimation technique in the nuclear community \cite{argaud_sensor_2018, argaud_stabilization_2017, gong_generalized_2022} highlighting the superior performance of SHRED. This paper presents the extension of the previous study to a parametric problem using the latest version of SHRED to show this methodology is general in scope, reliable, accurate and efficient for monitoring purposes and to assess the dynamics of quantities of interest in the whole domain under different conditions. Additionally, the performance of having mobile in-core sensors will be investigated, mimicking the possible scenario in which certain radionuclides, advected by the flow, can be traced through their emission. More in general, the nature of SHRED allows to tackle important issues in the nuclear community: the optimal configuration for sensors when some locations may be inaccessible, the indirect inference of unobservable fields and parametric datasets \cite{argaud_sensor_2018, ICAPP_plus2023, gong_parameter_2023}, all important \textit{desiderata} in fast, accurate and reliable \textit{digital twins} of the physical nuclear reactor \cite{mohanty_development_2021}, a topic of growing interest in the community.

The paper is structured as follows: at first, a brief presentation of the SHRED architecture is provided in Section \ref{sec: shred}; then, the MSFR and the case study for this work are discussed in Section \ref{sec: msfr}; Section \ref{sec: num-res} is devoted to the analysis of the main numerical results; finally, the main conclusions are drawn in Section \ref{sec: concl}.

\begin{figure*}[t]
    \centering
    \includegraphics[width=1\linewidth]{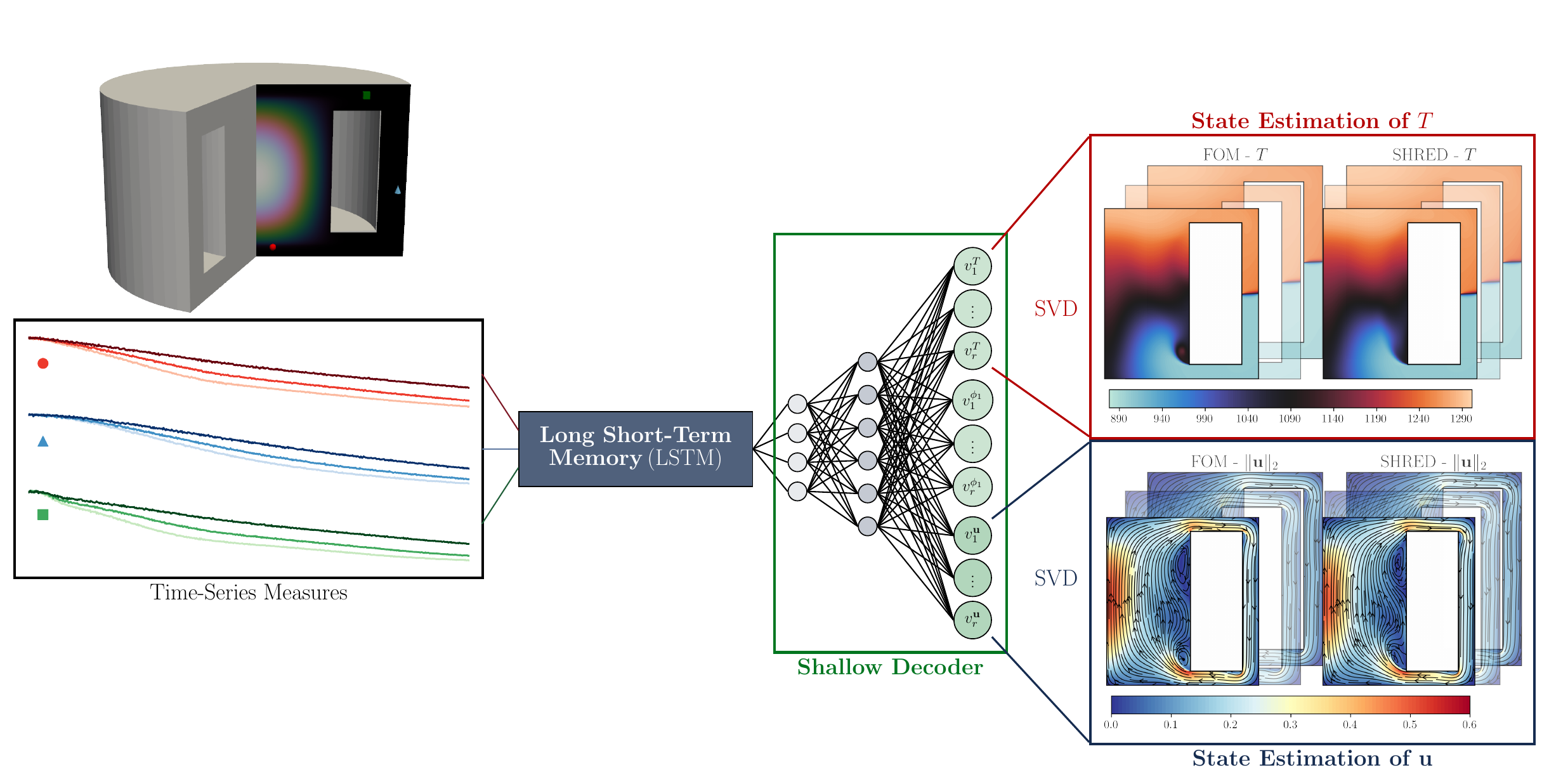}
    \caption{SHRED architecture applied to the Molten Salt Fast Reactor. Three out-of-core sensors are used to measure a single field variable $\phi_1$. The sensor time series are used to construct a latent temporal sequence model which is mapped to the compressive representations of all spatio-temporal field variables.  The compressive representations can then be mapped to the original state space by the singular value decomposition (SVD).}
    \label{fig: shred}
\end{figure*}

\section{SHallow REcurrent Decoder}\label{sec: shred}

The Shallow Recurrent Decoder is a novel neural network architecture \cite{williams2022data, ebers2023leveraging, kutz_shallow_2024} designed to map the trajectories of time-series measures $\vec{y}$ to a space spanned by some coefficients $\vec{v}$, which can be either high-dimensional or compressed through Singular Value Decomposition (SVD). Its basic version is composed of a Long Short-Term Memory (LSTM) network \cite{hochreiter1997long} and a Shallow Decoder Network (SDN) \cite{erichson2020shallow}. The combination of SVD-based reduction with this architecture is a good choice to generate surrogate models of physical systems \cite{kutz_shallow_2024, riva2024robuststateestimationpartial}: by compressing the starting dataset onto the latent space, the computational costs for the training phase are greatly reduced since SHRED has to learn the map from the trajectories of the measurements to a compressed space. Therefore, this work adopts the compressed version of SHRED, leveraging the SVD to retrieve the surrogate representation of the input data \cite{brunton_data-driven_2022}. Figure \ref{fig: shred} highlights the main structure of the SHRED network: first, the LSTM learns the temporal dynamics of the different trajectories according to Takens embedding theory, which mathematically guarantees that the sensor trajectory information of a single field is a diffeomorphic representation of all other fields \cite{takens1981lnm}; then, the SDN projects the hidden space back to the compressed space spanned by the SVD, to be later decompressed using the spatial modes. 

SHRED comes with important advantages compared to other data-driven ROM methods. First and foremost, the number of input sensors can be as low as 3 \cite{riva2024robuststateestimationpartial, kutz_shallow_2024, ebers2023leveraging}: previous works have shown that the reconstruction errors reach a plateau when this number is reached, meaning that even by increasing the number of sensors, no significant improvement in the performance is observed \cite{shredrom}. Additionally, SHRED can easily tackle multi-physics data starting from a single observable field, especially for strongly-coupled systems \cite{riva2024robuststateestimationpartial}: this is once again related to the Takens embedding theorem, making SHRED able to learn the non-linear dynamics between all quantities of interest, both observables and non-observables. The same holds for different domains: given an observable on a part of the domain, SHRED can recover both observable and unobservable fields in all regions of the domain, even sensor-blind ones.

Compared to other ML methods, training in the SVD-based SHRED occurs on the compressed data, thus enabling laptop-level and offline training \cite{kutz_shallow_2024}; additionally, SHRED requires minimal hyperparameter tuning, as demonstrated by its application on vastly different problems \cite{williams2022data, shredrom, riva2024robuststateestimationpartial, ebers2023leveraging}. The structure of the network architecture, in terms of layers and neurons, is reported in Table \ref{tab: SHRED-architecture}: it is worth mentioning that this is the exact same structure as the one of \cite{williams2022data, riva2024robuststateestimationpartial, shredrom}.

\begin{table}[t]
    \centering
    \begin{tabular}{c|c|c}
        \toprule $\,$ & Hidden Layer 1 & Hidden Layer 2 \\ \midrule
        LSTM & 64 & 64 \\ \midrule
        SDN & 350 & 400  \\ \bottomrule
    \end{tabular}
    \caption{Architecture of the SHRED networks.}
    \label{tab: SHRED-architecture}
\end{table}

Another significant feature of SHRED, which deserves separate discussion, is its agnosticism against sensor placement. Whereas most data-driven and ML methods require a (often computationally expensive) optimisation of the position of sensors, especially for safety-critical applications \cite{argaud_sensor_2018, gong_reactor_2024}, SHRED can retrieve the full state given three randomly placed sensors, intuitively following the principle of triangulation used in GPS and the fact that the previous sensor history is not discarded; instead, it is used to learn the temporal dynamics. The only two caveats in this agnosticism are that 1) the sensor must capture some dynamics (i.e., it should not record constant fields) and 2) the selected sensors must not all be aligned with each other.

The SHRED architecture has been implemented in Python using the PyTorch package \cite{pytorch}; the original code \cite{williams2022data} has been adapted for the application of this paper, and it is openly available under the MIT license at \textit{https://github.com/ERMETE-Lab/NuSHRED}.

Firstly, the SHRED was implemented in a single parameter configuration \cite{williams2022data, ebers2023leveraging, riva2024robuststateestimationpartial, kutz_shallow_2024}; the same architecture can be easily extended to parametric datasets \cite{shredrom} with minimal modifications compared to the standard version: in fact, the structure of SHRED is naturally conceived for the inclusion of multiple trajectories referring to different parameters, as the input data for the LSTM are lagged time-series data. The parameter $\boldsymbol{\mu}$ can then be added to the architecture, both as input (if known) or output if an estimation is needed. 

The most critical part of extending SHRED to parametric datasets is the data compression with the SVD. Given a snapshot matrix $\mathbb{X}^{\boldsymbol{\mu}_p}\in\mathbb{R}^{\mathcal{N}_h\times N_t}$ for a specific parameter $\boldsymbol{\mu}_p$, with $\mathcal{N}_h$ the spatial degree of freedom (i.e., the mesh size) and $N_t$ the saved time steps, the SVD allows to generate a basis $\mathbb{U}^{\boldsymbol{\mu}_p}\in\mathbb{R}^{\mathcal{N}_h\times r}$ of rank $r$ such that a latent representation $\mathbb{V}^{\boldsymbol{\mu}_p}=\left(\mathbb{U}^{\boldsymbol{\mu}_p}\right)^T\mathbb{X}^{\boldsymbol{\mu}_p}\in\mathbb{R}^{r\times N_t}$ can be obtained for that specific parameter. These coefficients $\mathbb{V}^{\boldsymbol{\mu}_p}$ embed the temporal dynamics and are used to train the SHRED; however, for a parametric dataset, it is necessary to obtain a common basis spanning the whole parametric space, thus encoding the underlying physics of the problem. If the dimension of the problem is sufficiently small, stacking the snapshots of the whole parametric dataset \rev{(being $N_p$ the number of parameters collected)} as 
\begin{equation}
    \mathbb{X} = \left[\mathbb{X}^{\boldsymbol{\mu}_1}|\mathbb{X}^{\boldsymbol{\mu}_2}| \dots | \mathbb{X}^{\boldsymbol{\mu}_{N_p}}\right]\in\mathbb{R}^{\mathcal{N}_h\times N_t\cdot N_p}
\end{equation} 
is the easiest way of proceeding: this option is feasible if the resulting matrix $\mathbb{X}$ fits the RAM of the machine that will run the training phase. Otherwise, hierarchical or incremental versions of the SVD on the starting, non-stacked dataset are necessary \cite{iwen_distributed_2016}. In both cases, the randomised version of SVD \cite{halko_finding_2010, bach_randomized_2019} is recommended for compression, as it has significant cost savings compared to the standard SVD since it works with batches from the original matrix. Given the dataset dimension of the problem considered in this paper, an investigation of incremental and hierarchical methodologies for SVD has been carried out in \ref{appendix}.

\subsection{Ensemble Strategy for Uncertainty Quantification}\label{sec: ensemble-shred}

\begin{figure*}[t]
    \centering
    \includegraphics[width=1\linewidth]{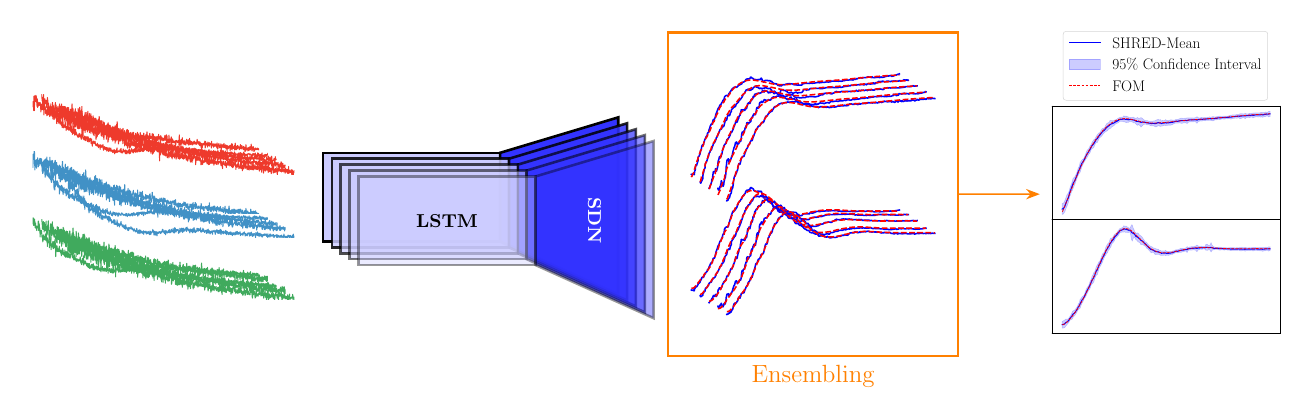}
    \caption{Scheme of the ensemble strategy for SHRED.}
    \label{fig: ensemble-shred}
\end{figure*}

As said above, SHRED is \textbf{agnostic} to sensor positions, meaning that this architecture completely decouples the problem of sensor positioning from that of state estimation in a data assimilation context, making it very flexible. The ability to use only a few randomly selected sensors for reconstructing the entire dynamics of a physical system can be exploited to improve the performance of SHRED, especially in the presence of noisy data \cite{riva2024robuststateestimationpartial}, without the need to increase the number of sensors used. This feature is attractive for nuclear reactors, in which the number of sensors must be kept as low as reasonably possible \cite{argaud_sensor_2018, gong_reactor_2024} to avoid perturbing the physical system \cite{argaud_sensor_2018} and for economical reasons, due to the harsh environmental conditions which significantly impact the useful lifetime of the sensor. 

In practice, suppose to have 10 available sensors: from these, different subsets (called configurations in the following) can be obtained: from statistics, it can be proved that the number of possible configurations is $\binom{10}{3}=120$; accordingly, since SHRED is quite cheap to be trained, it may be smart to avoid using all the sensors to train a single SHRED model, by instead sampling different configurations/subsets to train different SHRED models and then ensemble them to obtain a more robust state estimation \cite{riva2024robuststateestimationpartial}. In this way, the final prediction would be more statistically significant, as it will be characterised by a mean value and a standard deviation. Furthermore, this framework could detect possible sensor failures and eventual issues in the nuclear reactor by evaluating the uncertainty and comparing the different SHRED outputs: further studies regarding this topic are currently ongoing.

The ensemble strategy for SHRED is depicted in Figure \ref{fig: ensemble-shred}: the sparse measurements are collected and lagged; to predict the state at time $t_k$, the previous $k$ time instants are saved. For each sensor configuration $(l)$, the trajectories of the sensors are lagged as $\left[\vec{y}^{\boldsymbol{\mu}_i, (l)}_{k}, \dots , \vec{y}^{\boldsymbol{\mu}_i, (l)}_{k-K}\right]$ for $l=1, \dots, L$; the SHRED models are trained, producing $L$ outputs for the SVD coefficients $\mathbb{V}^{\boldsymbol{\mu}_i, (l)}$. The prediction of each SHRED model can be treated as a random variable represented as the sum of two contributions, a mean $\mathbb{V}^{\boldsymbol{\mu}_i}$ (being the generally unknown truth) and some disturbances ${\Sigma}_{(l)}^{\boldsymbol{\mu}_i}$ such that $\mathbb{V}^{\boldsymbol{\mu}_i, (l)} = \mathbb{V}^{\boldsymbol{\mu}_i}+{\Sigma}_{(l)}^{\boldsymbol{\mu}_i}$. The ensemble mode for $L$ SHRED models allows to produce a prediction $\hat{\mathbb{V}}^{\boldsymbol{\mu}_i}_L$ by averaging the different SHRED models, i.e.
\begin{equation}
    \hat{\mathbb{V}}^{\boldsymbol{\mu}_i}_L  = \frac{1}{L}\sum_{l=1}^{L}\mathbb{V}^{\boldsymbol{\mu}_i, (l)} =  \mathbb{V}^{\boldsymbol{\mu}_i}+\frac{1}{L}\sum_{l=1}^{L}{\Sigma}_{(l)}^{\boldsymbol{\mu}_i}
\end{equation}
which tends to the truth $\mathbb{V}^{\boldsymbol{\mu}_i}$ for $L\rightarrow +\infty$. The disturbances ${\Sigma}_{(l)}^{\boldsymbol{\mu}_i}$ are uncorrelated zero-mean random variables; this accounts for the variances introduced by the sensor positions, the noise of the input measurements, and any uncertainty related to the training dataset. This quantity can be estimated using the sample variance estimator
\begin{equation}
    \hat{\Sigma}_{\mathbb{V},L}^{\boldsymbol{\mu}_i} = \sqrt{\frac{1}{L-1}\sum_{l=1}^{L} \left(\mathbb{V}^{\boldsymbol{\mu}_i, (l)} - \hat{\mathbb{V}}^{\boldsymbol{\mu}_i}\right)^2}
\end{equation}

Therefore, the variance associated with the prediction can be directly computed as
\begin{equation}
    \text{Var}\left[\hat{\mathbb{V}}^{\boldsymbol{\mu}_i}_L\right] =\xi_L^2\approx \frac{\left(\hat{\Sigma}_{\mathbb{V},L}^{\boldsymbol{\mu}_i}\right)^2}{L}
    \label{eqn: ensemble-std}
\end{equation}

In this way, instead of having a single prediction, the ensemble strategy provides a more robust estimation of the state space, which is particularly useful in the presence of noisy data, allowing to improve the performance of SHRED and obtain more reliable results \cite{riva2024robuststateestimationpartial}.

\section{The Molten Salt Fast Reactor}\label{sec: msfr} 

Conventional nuclear reactors are characterised by solid fuel, usually in the form of uranium dioxide, a coolant, and a moderator which keeps the temperature under control and slows down the neutrons to enhance thermal fission events, respectively \cite{DuderstadtHamilton}. For the next generation of nuclear reactors, to improve economy, safety, security, and non-proliferation resistance, the Generation IV International Forum \cite{GenIV-RoadMap} has listed several innovative concepts, considering different coolants, fast energy spectrum for neutrons, and circulating fuel.

\begin{figure}[t]
  \centering
  \includegraphics[width=0.6\linewidth]{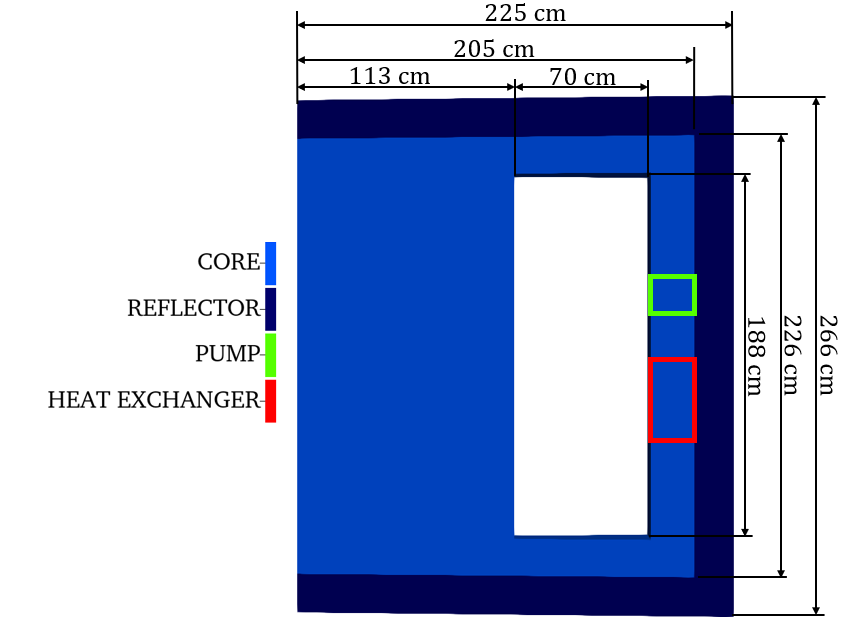}
  \caption{OpenFOAM simulation domain with the main geometric dimensions and the primary loop components. The geometry refers to a 2D axisymmetric wedge of the EVOL geometry of the European MSFR design, and includes molten salt fuel (light blue), the Hastelloy reflector (dark blue), the primary pump (green) and the heat exchanger with the intermediate cycle (red). The blank hole represents the solid salt fertile blanket, not simulated in the present model.}
  \label{fig: evol-geom}
\end{figure}

Among these, the Molten Salt Fast Reactor (MSFR) was selected as the reference concept for circulating fuel reactors, and it has been extensively studied within the EVOL, SAMOFAR, and SAMOSAFER projects \cite{brovchenko2013optimization}. This innovative design features a liquid fuel salt, composed of an eutectic mixture of $^7$LiF (77.5 mol\%) and $^{232}$ThF$_4$ (22.5 mol\%) combined with other heavy fluorides. This reactor is a unique example of a fast reactor, where the fuel is in a liquid state and homogeneously mixed with the coolant: this fact makes in-core sensing a challenging task due to the lack of solid structures, high fluences, high temperatures, and corrosion issues \cite{ICAPP_plus2023,riva_stefano_neutron_2024}. From a chemistry perspective, being the fuel dissolved into a high-temperature molten salt, actinide solubility, oxidation states and fission product behaviour all affect the operation of the system; additionally, understanding the thermo-chemical properties and behaviour of the salt is of paramount importance for fuel recycling, separation of fission products, and corrosion mitigation. This motivates the investigation of innovative sensing strategies to monitor the behaviour and the status of the reactor using out-of-core measurements or non-conventional in-core sensing, such as mobile probes, to, for example, track a specific radionuclide based on its emission. It is worth mentioning that the present work is not concerned with a feasibility study of this sensing solution: rather, it focuses on how information coming from such sensors can be used for state reconstruction.

Regardless of the selected sensing strategy, the SHRED architecture will adopt sparse and randomly placed sensors (either fixed or mobile) to reconstruct the entire state of the system in time under different conditions for the same accidental scenario. As a test case, this work adopts the 2D axisymmetric wedge (5$^o$) of the EVOL geometry of the MSFR \cite{brovchenko2013optimization} (Figure \ref{fig: evol-geom}), including also an additional external layer of thickness 20 cm to mimic the presence of the Hastelloy reflector \cite{PHYSOR24_MSFR_outcore, riva2024robuststateestimationpartial}, in which fixed sensors will be restricted. 

The mathematical description for this reactor is represented by a multiphysics model, using multi-group diffusion for neutronics and the incompressible single-phase version of the Reynolds-Averaged Navier-Stokes equations (RANS) with the Realizable $\kappa-\varepsilon$ turbulence model and the Boussinesq approximation to consider buoyancy effects (under the assumption that thermo-physical properties are constant within the same region)for thermal-hydraulics \cite{aufiero2014development, CERVI2019379}. The model has been implemented and solved in OpenFOAM \cite{weller_tensorial_1998}. This reactor is particularly interesting because of the intrinsic multiphysics and strong coupling between neutronics and thermal-hydraulics, making it a good candidate to show the indirect reconstruction capabilities of state estimation methods. The model adopted has been verified within the European framework for MSFR against other codes, \cite{TIBERGA2020107428}; the multiphysics solver was instead verified and validated in \cite{aufiero2014tesi}. In this work, only synthetic data will be adopted since, at the current stage, there are no experimental data available for this particular reactor, leaving the use of SHRED on real facilities a matter for future studies \cite{NURETHDynasty}. A summary of the governing equations is discussed in \ref{app: msfr-gov-eqn}.

The accidental scenario considered in this work is the Unprotected Loss of Fuel Flow (ULOFF), in which the flow rate of the pump is decreased exponentially $\sim e^{-t/\tau}$, resulting in a decrease of the velocity magnitude inside the reactor, affecting the power-to-flow ratio. Different values of $\tau$, specifically $N_p=21$, have been considered within the range $[1, 10]\,\text{s}$ and each case is simulated for 30 seconds, with a saving time of 0.05 s, resulting in $N_t=600$ snapshots for each instance of the parameter $\tau$. The number of parameters was chosen to have a good balance between the computational time to run a single-parameter FOM instance and a reasonable number of parameters (indeed, evaluation of the completeness of the training dataset remains a matter of expertise). A single FOM simulation takes about 8 hours of wall-clock time using 5 CPUs on a high-performance cluster (Intel Xeon E5-2640-V4 @ 128 GB RAM).

Several fields are needed to fully describe the neutron economy and the thermal-hydraulics of the system: in particular, for this case, six energy-group fluxes $\{\phi_g\}_{g=1}^6$, eight groups of delayed neutrons $\{c_k\}_{k=1}^8$, the total flux $\Phi$ and the power density $q'''$ are considered for the neutronic side, to which the thermal-hydraulics fields, namely pressure, temperature, velocity and turbulent quantities $(p, p_{rgh}, T, \vec{u}, \kappa, \nu_t)$, must be added. Except for the velocity $\vec{u}$, all the others are scalar fields. Overall, the full-order state space vector $\mathcal{V}$ is represented by 22 different coupled fields, i.e. 
\begin{equation}
    \mathcal{V} = \left[\phi_1, \dots, \phi_6, c_1, \dots, c_8, \Phi, q''', p, p_{rgh}, T, \vec{u}, \kappa, \nu_t \right] 
    \label{eqn: state-space}
\end{equation}

Since in real engineering systems it is not always possible to have access to all quantities of interest \cite{gong_data-enabled_2022, INTROINI2023109538, argaud_sensor_2018}, the SHRED architecture will be used to reconstruct all quantities of interest starting from the measurements of only one observable field; this is made possible because the MSFR (and more in general, nuclear reactors) is a strongly coupled system, where each field carries some information about the other quantities \cite{INTROINI2023109538}. In particular, for fixed sensor the logic is to use easy-to-measure quantities (temperature, fluxes) to reconstruct hard-to-measure ones (velocity, precursor concentrations).

\begin{figure*}[tp]
    \centering
    \includegraphics[width=1\linewidth]{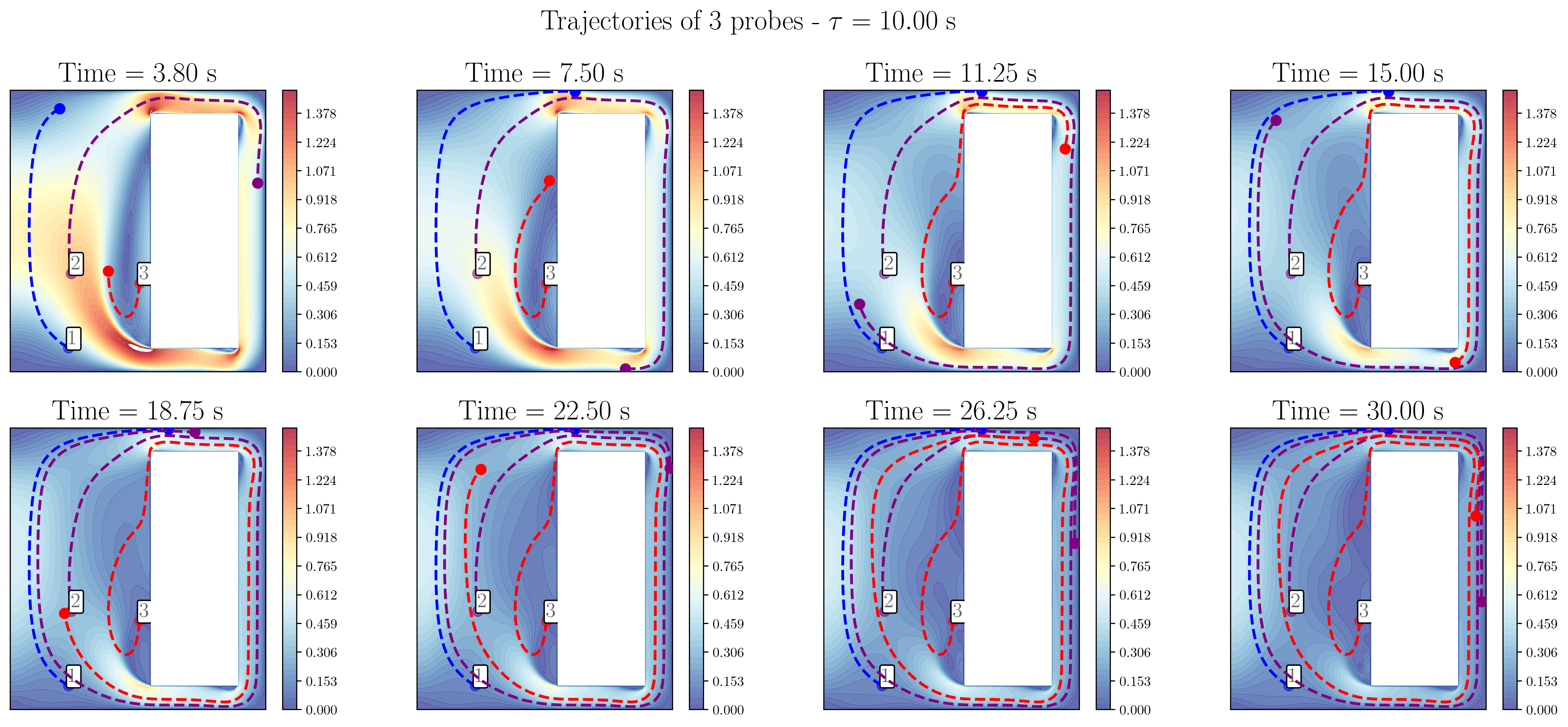}
    \caption{Example of trajectories of 4 probes inside the liquid core over time at the test parameter $\tau=10.00$ seconds. Starting from the bottom left part of the reactor, the probes are advected by the velocity field in a pure Lagrangian way.}
    \label{fig: probe-traj}
\end{figure*}

As already mentioned, this work considers different sensing strategies for the MSFR from the theoretical point of view, that is, regardless of their practical feasibility (especially for what concerns mobile sensors); thus three different sensing strategies will be considered to reconstruct the entire state space $\mathcal{V}$:
\begin{enumerate}
    \item \textbf{Out-of-core sensors}: sensors are placed in the reflector region, outside the core, and they measure the fast flux $\phi_1$ (the only observable field). The possibility of measuring different fields as well, reflecting the actual availability of sensors, will be investigated in future works; as a rule of thumb, measured fields are typically quantities easy to obtain, like fluxes or temperature. It is much harder to have access to quantities like the velocity field. 
    \item \textbf{Mobile sensors}: the first group of precursors is measured along the trajectory of a particle advected by the flow. Both the concentration of the precursors and the trajectory represent the measurements in this case: this case simulated the possibility of tracking the emission of a particular radionuclide within the core.
    \item \textbf{Mobile probes}: inert particles within the liquid core are tracked, and only their positions are collected over time (akin to a buoy).
\end{enumerate}

Especially for fixed sensors, the fact that SHRED is agnostic to sensor positions \cite{williams2022data, kutz_shallow_2024, riva2024robuststateestimationpartial} allows for placing sensors in the available regions of the nuclear reactor without losing performance, which is generally not true for other methods like the the Generalised Empirical Interpolation Method \cite{argaud_sensor_2018, ICAPP_plus2023}, in which there is a deterioration of the \textit{a-priori} error estimation due to the constraints on the sensor positions \cite{maday_convergence_2016}.

Figure \ref{fig: probe-traj} shows the trajectories of 3 probes inside the liquid core starting from the bottom region of the core domain; from these trajectories, an important issue related to mobile probes is the deposition on the boundary of the tracked particle \cite{DIRONCO2022111739} (as for particle 1 in Figure \ref{fig: probe-traj}). This fact may affect the performance of SHRED because the signal for this particle will be constant after a certain time instant; nevertheless, the ensemble strategy could help in identifying these particles, ensuring, up to a certain point, some self-correction capabilities. More developments on this task are foreseen in future works.

As briefly mentioned, the particles for the last two cases will be considered point-wise, and they are assumed to be purely advected by the flow velocity $\vec{u}$, without any decay or interaction with the other fields. A Lagrangian particle tracking approach \cite{versteeg2007introduction} is used to track the position of the particles: let $\vec{s}(t)\in\mathbb{R}^2$ be the position of a particle at time $t$, the equation of motion is given by
\begin{equation}
    \frac{d\vec{s}(t)}{dt} = \vec{u}(\vec{s}(t), t) \qquad \vec{s}(0) = \vec{s}_0.
\end{equation}
This equation has been discretised with an Implicit Euler scheme and solved in Python.

\section{Numerical Results}\label{sec: num-res}

The dataset adopted for this work consists of $N_p=21$ simulations for different values of $\tau$ within the range $[1,10]$. The dataset has been divided into train (71.4\%-15 parameters), test (14.3\%-3 parameters), and validation (14.3\%-3 parameters) using random splitting: both parametric interpolation and extrapolation are considered, whereas only temporal reconstruction is investigated, without considering any forecasting capabilities for the time being. The number of time instances $N_t$ is 600, whereas the mesh is composed of $\mathcal{N}_h=94611$ hexahedron cells. The training dataset is used to generate the SVD basis and to train the SHRED architecture. Focusing on the first step, the snapshots of each field in $\mathcal{V}$ have been organised into stacked matrices, as discussed in Section \ref{sec: shred}. Due to the different magnitudes of the fields, the snapshots have been normalised as follows: the generic field $\psi$ is rescaled against its minimum and maximum value in critical conditions (the initial steady condition, which is common for the whole dataset), so that its range is always $[0,1]$:
\begin{equation}
    \mathbb{X}_{\psi, ij}^{\boldsymbol{\mu}_p} \longleftarrow \frac{ \psi(\vec{x}_i; t_j, \boldsymbol{\mu}_p) - \min\limits_{\vec{x}\in\Omega}\psi(\vec{x}; 0)}{\max\limits_{\vec{x}\in\Omega}\psi(\vec{x}; 0) - \min\limits_{\vec{x}\in\Omega}\psi(\vec{x}; 0)}
    \label{eqn: rescaling}
\end{equation}

Then, the randomised SVD is applied on the stacked matrix $$\mathbb{X}_\psi = \left[\mathbb{X}_\psi^{\boldsymbol{\mu}_1}|\mathbb{X}_\psi^{\boldsymbol{\mu}_2}| \dots | \mathbb{X}_\psi^{\boldsymbol{\mu}_{N_p}}\right]$$, retrieving a reduced representation of each field in terms of the first $r$ principal components. By looking at the decay of the singular values \cite{rozza_model_2020}, $r$ is taken to be 10 for all the fields, ensuring that at least 99\% of the total information is encoded into the basis functions.

\begin{figure*}[tp]
    \centering
    \includegraphics[width=1\linewidth]{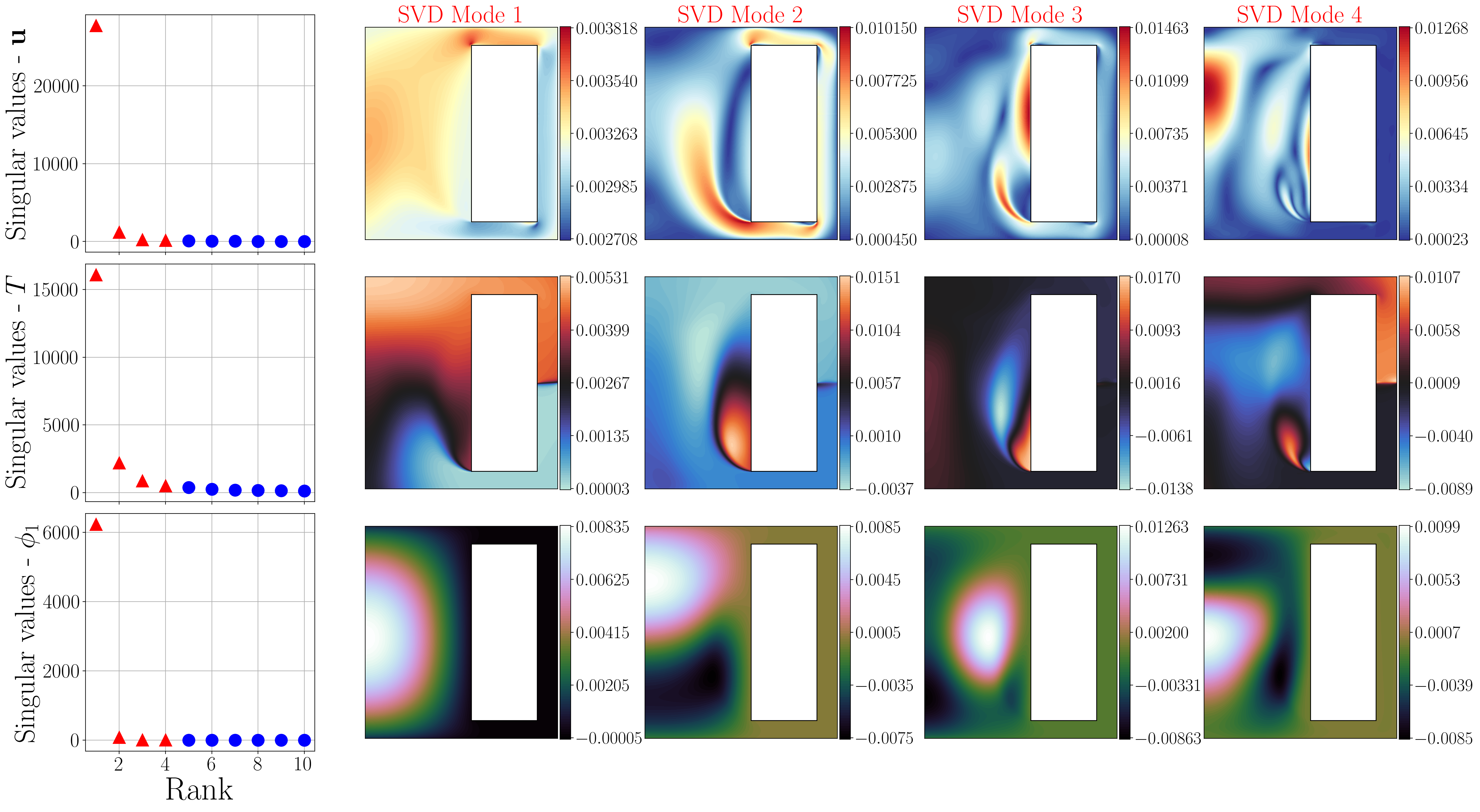}
    \caption{Decay of the singular values and contour plots of the first 5 SVD modes of velocity $\mathbf{u}$, temperature $T$ and fast flux $\phi_1$, underlining the hierarchical spatial features. The singular values for all the three quantities show an exponential decay, highlighting the fact that low rank modes are the most important.}
    \label{fig: svd-modes}
\end{figure*}

The decay of the singular values for velocity $\mathbf{u}$, temperature $T$, and fast flux $\phi_1$, and the contour plots of the first four modes are displayed in Figure \ref{fig: svd-modes}. Each singular value from the SVD denotes the amount of information contained in the associated spatial mode \cite{brunton_data-driven_2022}: a fast decay means that the majority of information is contained in the first few modes, which therefore contain the key spatial dynamics of the system. It is visible how the first mode is more dominant compared to the others, capturing the overall spatial dynamics of the data, whereas higher-order modes show lower and lower scales \cite{sirovich_turbulence_1987-1}. The singular values show an exponential decay, indicating how the first few modes embed the most important spatial features of the starting dataset and, overall, that the problem is indeed reducible.

Focusing on the input of SHRED, the fixed sensors can be placed only in the reflector region (Figure \ref{fig: evol-geom}) and their only observable field is the fast flux $\phi_1$; the initial position for mobile sensors has been chosen to be in the downcomer region (near the heat exchanger) and in addition to the position the concentration of the first group of precursors at the position $\vec{s}(t)$ can be measured; mobile probes can randomly start in the bottom region of the reactor to see how constraints on the initial position affects mobile sensing: in the current design, helium bubbles will be injected from the bottom right corner of the core, for fission product removal and potentially reactivity control, thus the selection of this region for the possible starting position of sensors/probes. For the out-core sensor case, the input dimension for SHRED is $s=3$; for a single SHRED architecture, the mobile sensor follows a single particle, tracking its position $\vec{s}(t)\in\mathbb{R}^2$ and measuring the concentration of the first group of precursors $c_1$, thus the input dimension of SHRED remains $s=3$; for the case with mobile probes, where only the positions of the particles is tracked, three probes, each following one particle, are considered, thus having an input dimension of $s=6$ (this solution, instead of following one or two particles at a time, has been proved to be the most effective according to the numerical experience of the authors). 

Let $\vec{y}\in\mathbb{R}^s$ be the input vector for SHRED (the parameter has been omitted for the sake of notation simplicity), including either the local evaluation of the quantity of interest (the fast flux or the precursor concentration) or the trajectory of the particle: the local evaluations of either the fast flux or precursors have been synthetically generated from the OpenFOAM data, assuming the sensor to be point-wise (an extension to local averages can be found in \cite{argaud_sensor_2018, ICAPP_plus2023, riva2024robuststateestimationpartial}), and polluted with Gaussian noise $\epsilon\sim \mathcal{N}(0,\sigma^2)$. The measure associated to the generic field $\psi$ (normalised) for the sensor in position $\vec{s}_k(t)$ (either fixed or mobile) is defined as 
\begin{equation}
    y_k^{\psi}(\cdot) = (1+\epsilon)\cdot  \int_\Omega \tilde{\psi}(\vec{x}; \cdot) \cdot \delta (\vec{x} - \vec{s}_k(t))\, d\vec{x}
    \label{eqn: sens-def}
\end{equation}
given $\delta$ the Dirac delta, representing the point-wise evaluation. Being more specific, for the fixed sensor case, the input vector SHRED is given by the observation of the fast flux in 3 points $[\vec{s}_1, \vec{s}_2, \vec{s}_3]$ in the reflector region, as
\begin{equation}
    \vec{y} = \left[ y_1^{\phi_1}, y_2^{\phi_1}, y_3^{\phi_1}\right] \in \mathbb{R}^3  
\end{equation}
where $ y_k^{\phi_1}$ is the measurements at the coordinates of the sensors $[\vec{s}_1, \vec{s}_2, \vec{s}_3]$, defined as in Eq. \eqref{eqn: sens-def}. For the mobile sensor case, the input vector is the concentration of the first group of precursors at the position of the particle and its position:
\begin{equation}
    \vec{y} = \left[  y_1^{c_1}, s_{x,1}(t), s_{y,1}(t)\right] \in \mathbb{R}^3
\end{equation}
For the mobile probe case, the input vector is given by the position of the particles:
\begin{equation}
    \vec{y} = \left[ s_{x,1}(t), s_{y,1}(t), s_{x,2}(t), s_{y,2}(t), s_{x,3}(t), s_{y,3}(t)\right] \in \mathbb{R}^6
\end{equation} 

Previous analyses on noisy data have shown that SHRED is robust against random noise, particularly when used in ensemble mode \cite{riva2024robuststateestimationpartial, shredrom}. Since the SHRED architecture is quick to train \cite{williams2022data} and it does not require powerful GPUs, to the point that even personal computers can be used to perform the training phase, several SHRED models can be trained with different random positioned sensors to produce different outputs, i.e. different predictions of the reduced state space vector $\vec{v}$. From these, the sample mean and the associated standard deviation can be obtained (see Section \ref{sec: ensemble-shred}). 

In terms of computational costs, each SHRED model takes about 15 minutes of wall-clock time for the training phase on a personal computer with Intel Core i7-9800X CPU, whose clock speed is 3.80 GHz; to get a new output, the associated computational cost to obtain an estimation from a trained SHRED model is on the order of seconds.

\subsection{Reconstruction from out-core sensors}

In this first part, the reconstruction of the full state space from fixed out-core sensors, measuring the fast flux, is considered, following the work from the authors in \cite{riva_stefano_neutron_2024, ICAPP_plus2023}. At first, the latent dynamics spanned by the reduced coefficients $\vec{v}$ are plotted for the test parameters over time; then, the prediction given by SHRED is projected back to the full space and compared with the high-fidelity solution of the OpenFOAM model. For this case, the input vector $\vec{y}$ is polluted by random noise with standard deviation $\sigma=0.01$ as in \cite{riva2024robuststateestimationpartial}: preliminary investigations on the effect of noise have been performed showing that adopting the ensemble mode, the state can be accurately reconstructed; a more accurate analysis will be done in the future. Moreover, $L=10$ different sensor configurations have been used: the number of SHRED models $L$ adopted can be actually quite low and it generally depends on the number of sensors available, in this work $L$ has been taken equal to 10 following a sensitivity analysis on the uncertainties and theoretical calculations on the variance of the mean estimator (see Section \ref{sec: ensemble-shred}).

\begin{table}[htbp]
\centering
\begin{tabular}{c|c|c}
\textbf{\# Config.} & \textbf{Std. Dev. $\xi_L$} & \textbf{Norm. Std. Dev. $\xi_L/\xi_L^{\mathrm{max}}$} \\
\hline
2  & 0.0133 & 0.7790 \\
4  & 0.0093 & 0.5424 \\
6  & 0.0073 & 0.4253 \\
8  & 0.0079 & 0.4604 \\
10 & 0.0069 & 0.4060 \\
12 & 0.0070 & 0.4081 \\
14 & 0.0065 & 0.3818 \\
15 & 0.0067 & 0.3941 \\
20 & 0.0063 & 0.3709 \\
25 & 0.0058 & 0.3394 \\
30 & 0.0053 & 0.3109 \\
\end{tabular}
\caption{Standard deviation and normalized standard deviation of $\xi_L$ for selected numbers of configurations.}
\label{tab:sensitivity-ensemble}
\end{table}

As said, SHRED is used in ensemble mode, thus $L$ different models have been trained. The output of each model is used to get a mean and a standard deviation of the prediction. Let $\xi_L$ be the standard deviation of the mean estimator, from Eqn. \eqref{eqn: ensemble-std} (averaged with respect to parameters/time and size of the output of SHRED, i.e. reduced space), Table \ref{tab:sensitivity-ensemble} reports the results of a sensitivity analysis on the number of SHRED models $L$: it is important to notice that for $L>10$, the value of $\xi_{L}$ decreases less from $0.007$, whereas with respect to the worst case ($L=2$) the decrease is less dominant. This means that after 10 models, there is no significant improvement in the estimation of the uncertainty, making a good value for this problem.

\subsubsection{Learning the latent dynamics}

\begin{figure*}[tp]
    \centering
    \includegraphics[width=1\linewidth]{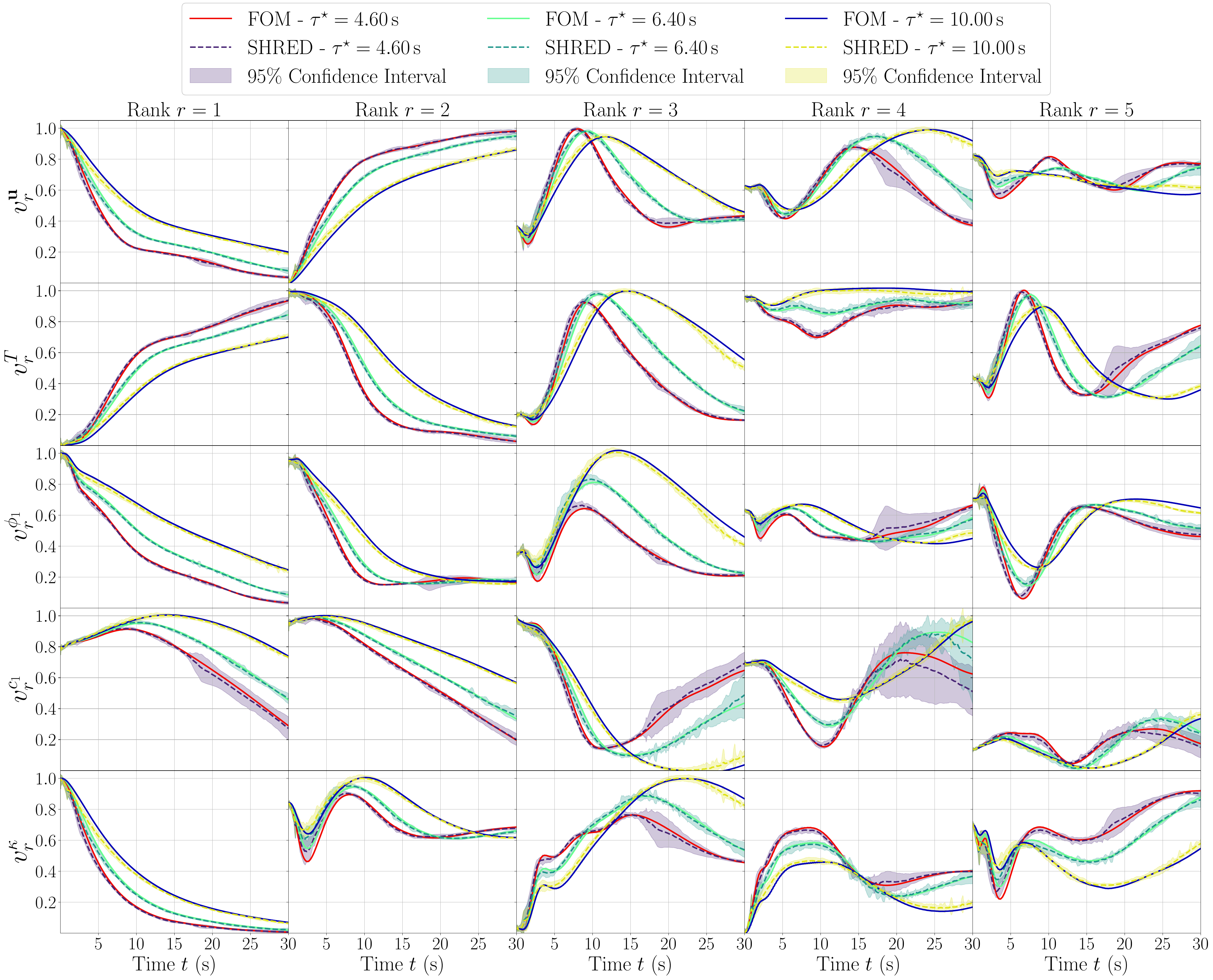}
    \caption{Comparison of the SHRED reconstruction, normalised to [0,1] of the first 5 SVD coefficients of velocity $\vec{u}$, temperature $T$, fast flux $\phi_1$ (observed field), first precursors group $c_1$ and turbulent kinetic energy $\kappa$,  for each test parameter. Dashed curves represent the mean of the SHRED models, the continuous lines are the ground truth (from the full-order data) and the shaded areas highlight the uncertainty regions for the SHRED models.}
    \label{fig: latent-dyn}
\end{figure*}

At first, the performance of SHRED for the state estimation during parametric accidental scenarios in the MSFR is assessed. Once all $L=10$ architectures have been trained, the output is compared with the full-order test dataset. Given $\hat{\mathbb{V}}^{\boldsymbol{\mu}_i}_L$ the ensembled SHRED prediction and $\mathbb{V}^{\boldsymbol{\mu}_i}$ be the associated truth (the reduced/latent dynamics) for test parameter $\boldsymbol{\mu}_i$, recalling that the rows represents the reduced coefficients and the columns the time instances. The overall SHRED test error on the output is defined as

\begin{equation}
    e = \frac{1}{N_s^{test}}\sum_{n=1}^{N_s^{test}}\left( \frac{1}{N_t}\sum_{j=1}^{N_t}\frac{\norma{\hat{\mathbb{V}}^{\boldsymbol{\mu}_n}_{L,j}- \mathbb{V}^{\boldsymbol{\mu}_n}_j}_2}{\norma{\mathbb{V}^{\boldsymbol{\mu}_n}_j}_2}\right)
\end{equation}
with the subscript $j$ indicating the corresponding column of the matrices. This is an average (to time and parameters) relative error (in Euclidean norm) measuring the overall performance of SHRED: for trained architecture, a value of 4.6\% is obtained, highlighting how the SHRED architecture can reconstruct the overall latent space. More specifically, Figure \ref{fig: latent-dyn} shows the reduced state space vector $\vec{v}$ of the modal SVD coefficients for the test parameters, normalised to $[0,1]$, for the velocity $\vec{u}$, the temperature $T$, the fast-flux $\phi_1$, the first group of precursors $c_1$ and the turbulent kinetic energy $\kappa$. There is an excellent agreement between the dashed curves representing the SHRED mean prediction and the ground truth from the starting dataset, especially for the lower ranks, which are the ones retaining most of the information content \cite{rozza_model_2020}. The SHRED architecture can map the trajectories of the sensor measurements almost correctly to the latent dynamics, thus retrieving the actual dynamics. Moreover, the advantage of fast training of different models allows for a more robust prediction, as it allows retrieving an uncertainty band, making the state estimation more reliable \cite{riva2024robuststateestimationpartial}.

\subsubsection{Decoding to the high-dimensional space}

Once the latent dynamics have been predicted by the SHRED architecture, it is possible to project the output of SHRED back to the high-dimensional space, using the SVD modes associated with each field. The results can be compared with the actual solution of the PDEs to assess if the accuracy is maintained at the full-order level. In particular, the average relative error $\varepsilon_2^\psi$ is defined as
\begin{equation}
    \varepsilon_2^{\psi} = \frac{1}{N_s^{test}}\sum_{n=1}^{N_s^{test}}\left( \frac{1}{N_t}\sum_{j=1}^{N_t}\frac{\norma{\vec{x}_{\psi,j}^{\boldsymbol{\mu}_n} - \hat{\vec{x}}_{\psi,j}^{\boldsymbol{\mu}_n}}_2}{\norma{\vec{x}_{\psi,j}^{\boldsymbol{\mu}_n}}_2}\right)
\end{equation}
in which $\vec{x}_{\psi,j}^{\boldsymbol{\mu}_n}\in\mathbb{R}^{\mathcal{N}_h}$ represents the true state vector for the generic field $\psi$ at time $t_j$ for the operating condition/parameter $\boldsymbol{\mu}_n$ and $\hat{\vec{x}}_{\psi,j}^{\boldsymbol{\mu}_n}$ is the correspondent SHRED prediction (averaged with respect to the configurations, see Section \ref{sec: ensemble-shred}).

\begin{figure}[tp]
    \centering
    \includegraphics[width=1\linewidth]{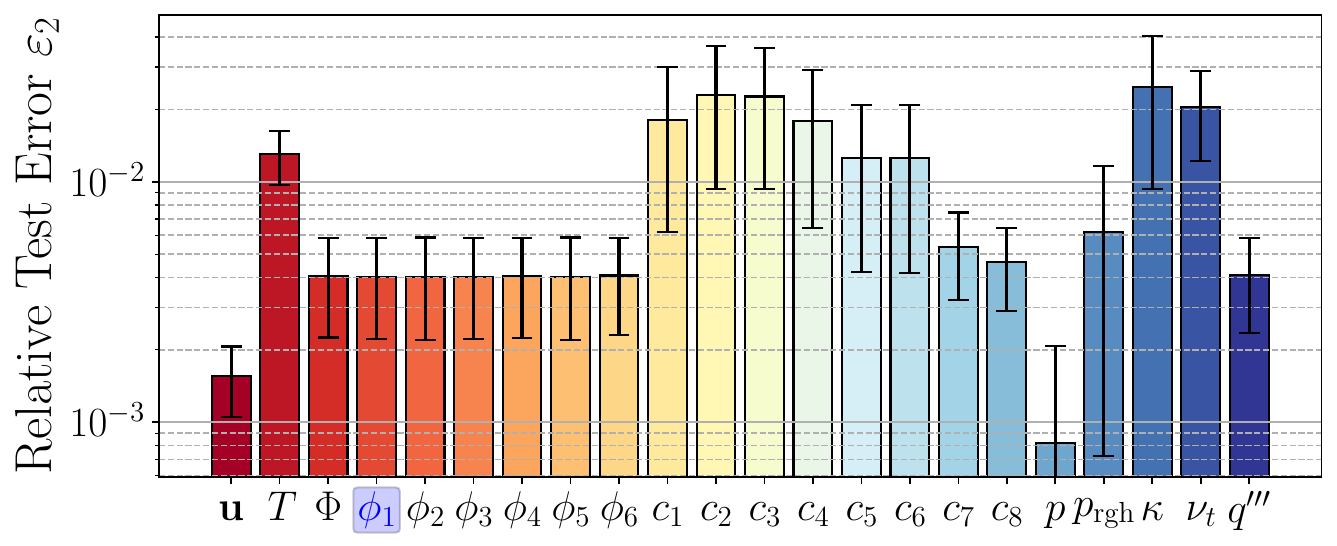}
    \caption{Average relative error measured in Euclidean norm (spatially) for all the reconstructed fields in the test set: each field presents a discrepancy lower than 2\%. In blue, the label of the measured field $\phi_1$ has been highlighted.}
    \label{fig: errorss}
\end{figure}

\begin{figure*}[tp]
    \centering
    \includegraphics[width=1\linewidth]{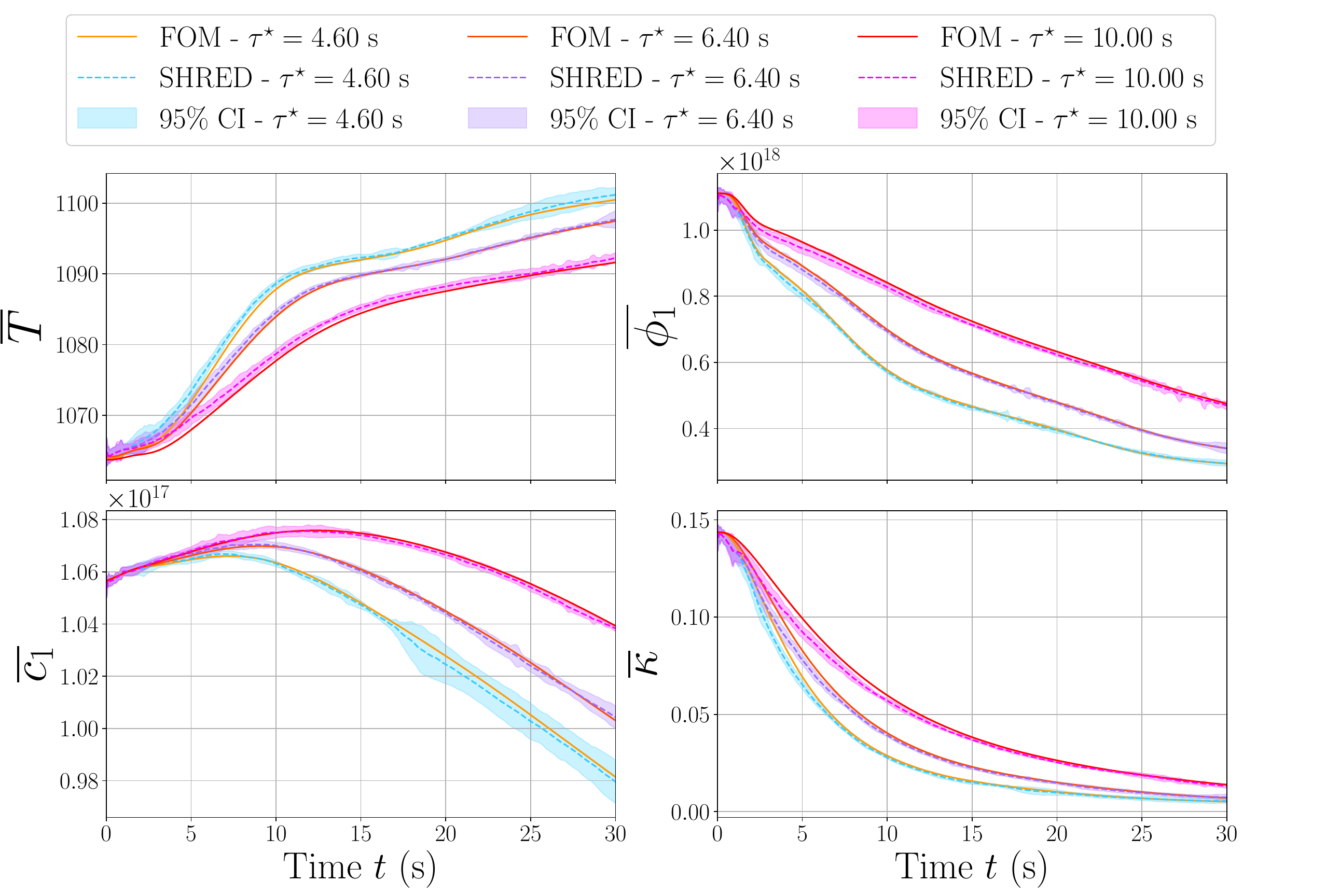}
    \caption{Comparison of the full-order and the SHRED prediction for the spatial average quantities of $T, \phi_1, c_1, \kappa$, with confidence interval 95\%, showing almost perfect agreement.}
    \label{fig: qoi-dynamics}
\end{figure*}

For this specific problem, the most difficult fields to reconstruct are the precursor concentrations and the turbulent quantities, whereas pretty good accuracy is obtained for temperature, power density, velocity and neutron fluxes: this is probably due to slower decay of the SVD singular values, which can be observed in the complementary Github repository, which are the quantities typically monitored in nuclear reactors to ensure the overall safety of the system (Figure \ref{fig: errorss}); nevertheless, over the test set the error is always below 2\% for all the fields. Figure \ref{fig: qoi-dynamics} shows instead the dynamics of the spatial average of temperature, total flux (directly connected to the power density), the first group of precursors, and turbulent kinetic energy: the accuracy of the SHRED prediction is extremely close to the full-order value, showing that this architecture is well suited for online monitoring.

\begin{figure*}[tp]
    \centering
        \begin{overpic}[width=0.95\linewidth]{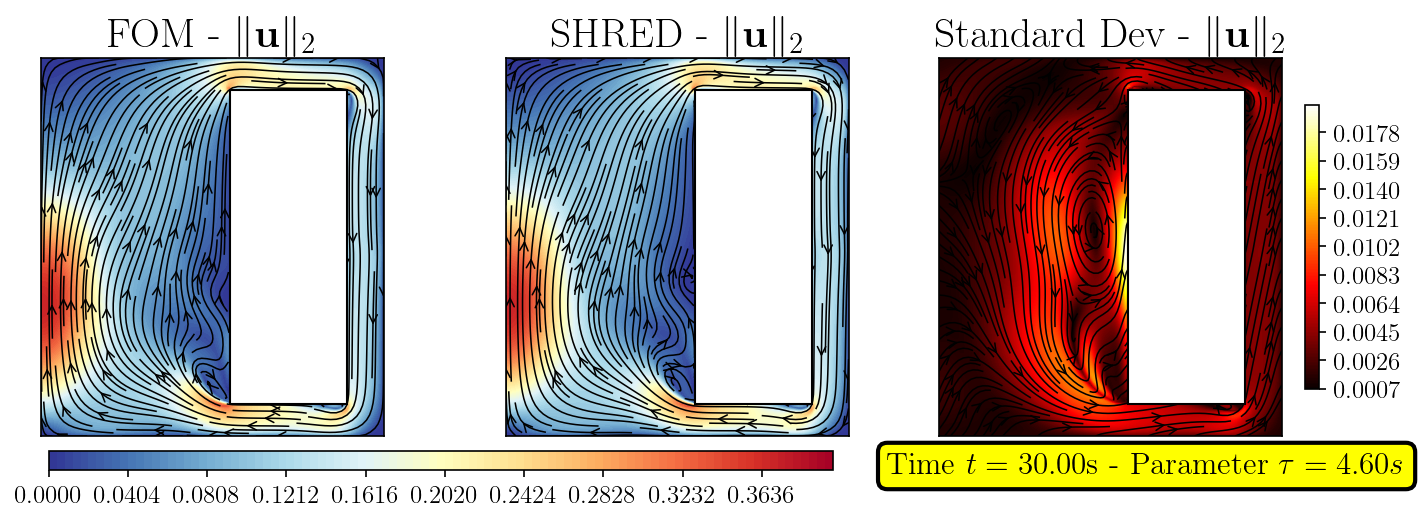}
        \put(-2.5,32){(a)}   
        \end{overpic}
        \begin{overpic}[width=0.95\linewidth]{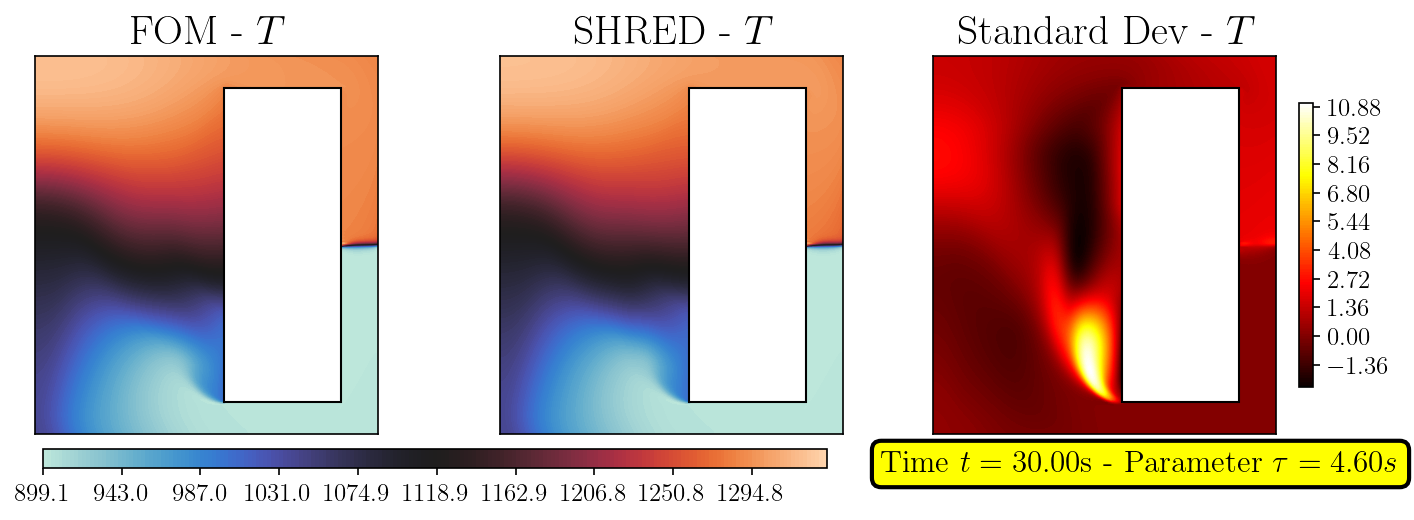}
        \put(-2.5,32){(b)} 
        \end{overpic}
        \begin{overpic}[width=0.95\linewidth]{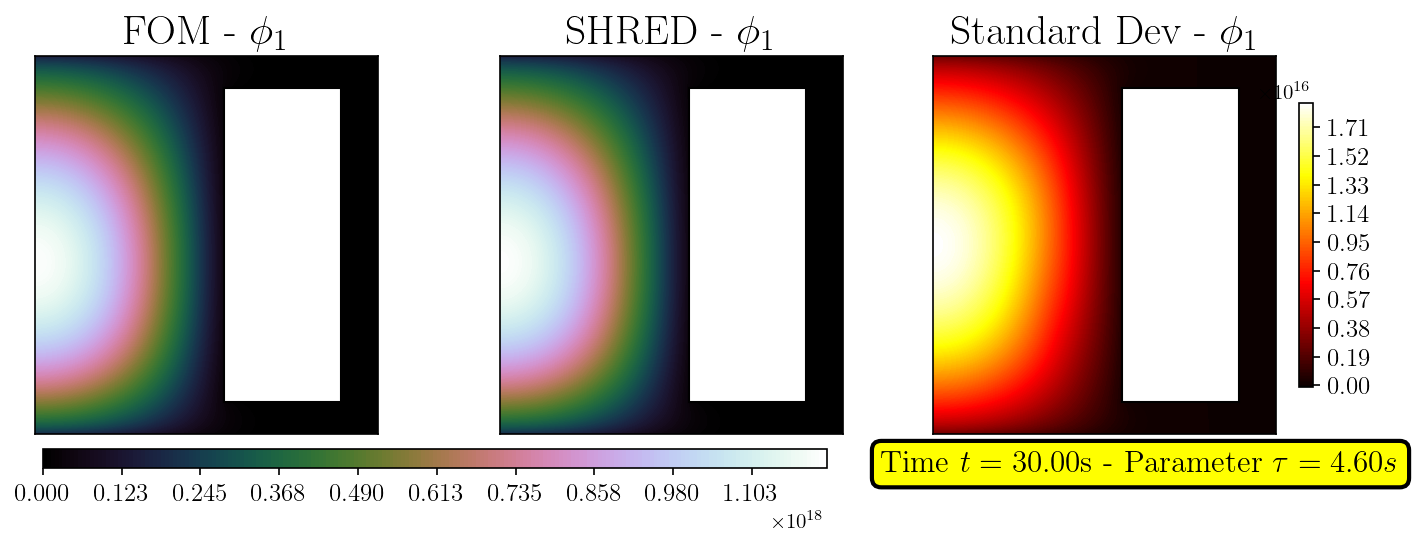}
        \put(-2.5,32){(c)} 
        \end{overpic}
    \caption{Contour plots at the last time step for the test parameter $\tau^*$ = 4.62 s of the velocity $\vec{u}$ (a), the temperature $T$ (b) and the observed field $\phi_1$ (c). From left to right: full-order solution, mean of the SHRED models and associated standard deviation. The prediction with SHRED provides a correct local state estimation both of the observable and the un-observable quantities; the right-most column, showing the standard deviation field, allows to see the locations with the most uncertainty, highlighting where the estimation is poorer.}
    \label{fig: contours}
\end{figure*}

Not only globally, but the SHRED architecture can reliably produce state estimation of the quantities of interest over the whole domain, which is one of the main advantages, along with their low computational cost, of ROM approaches compared to integral approaches. Figure \ref{fig: contours} shows some contour plots of the SHRED prediction at the last time step for the test parameter $\tau^\star = 4.6$ s for the velocity, the temperature, and the fast flux and their standard deviation.

The SHRED model can provide a correct state estimation of the observable field $\phi_1$ and the unobservable ones, such as temperature $T$ and velocity $\vec{u}$. In particular, the chosen rank of the SVD includes sufficient information to obtain a reliable local state estimation and to predict even some low-scale features of the SVD compression, especially for the velocity field: in fact, the remaining part of the recirculation region, near the bottom left corner of the blanket, is seen by the SHRED. If lower scale structures are present, which can play a significant role in highly turbulent flows (modelled with large eddy simulations), these can be estimated by increasing the rank to include low-scale features, or adopting some non-linear compression techniques as in \cite{Eiximeno2024}. Training more SHRED architectures also allows visually reporting the standard deviation of all quantities of interest, highlighting the regions where the uncertainty is higher (and therefore the SHRED reconstruction is poorer). Future works will focus on leveraging the SHRED results to develop greedy-like sensor positioning techniques.

Overall, the SHRED has been proven to be a strong and reliable tool in the state reconstruction problem of quantities of interest (both observable and unobservable) during parametric accidental scenarios. 

\subsection{Mobile Sensors and Probes}

In this section, the reconstruction capabilities of SHRED when measurements are taken from mobile sensors and probes, following their definition from Section \ref{sec: msfr}, are going to be discussed and compared with the fixed out-of-core sensors strategy presented before. In this work, the mobile sensors/probes strategy is investigated from a numerical point of view, without entering into the details on how to design suitable sensors that can address this task, acting more as a methodological study with broader applications outside the nuclear reactor framework. If suitable and well-performing, future research will focus on the practical application of these sensing strategies.

\begin{table}[tp]
    \centering
    \renewcommand{\arraystretch}{1.2}
    \begin{tabular}{|c|ccc|}
        \specialrule{1.2pt}{0pt}{0pt}  
        \multicolumn{4}{|c|}{\textbf{Average Relative Error} $\varepsilon_2$} \\
        \specialrule{1.2pt}{0pt}{0pt}  
        Field & Out-Core & Mobile Sensors & Mobile Probes \\
        \specialrule{1.2pt}{0pt}{0pt}  
        $\mathbf{u}$ & \cellcolor{green!20} 0.001558 & 0.003515 & \cellcolor{red!20} 0.004102 \\
        $T$ & \cellcolor{green!20} 0.013003 & 0.017279 & \cellcolor{red!20} 0.020116 \\
        $\Phi$ & \cellcolor{green!20} 0.004057 & 0.023735 & \cellcolor{red!20} 0.039459 \\
        $\phi_1$ & \cellcolor{green!20} 0.004025 & 0.023763 & \cellcolor{red!20} 0.039601 \\
        $\phi_2$ & \cellcolor{green!20} 0.004032 & 0.023687 & \cellcolor{red!20} 0.039527 \\
        $\phi_3$ & \cellcolor{green!20} 0.004036 & 0.023771 & \cellcolor{red!20} 0.039503 \\
        $\phi_4$ & \cellcolor{green!20} 0.004044 & 0.023745 & \cellcolor{red!20} 0.039651 \\
        $\phi_5$ & \cellcolor{green!20} 0.004035 & 0.023753 & \cellcolor{red!20} 0.039464 \\
        $\phi_6$ & \cellcolor{green!20} 0.004078 & 0.023785 & \cellcolor{red!20} 0.039595 \\
        $c_1$ & \cellcolor{green!20} 0.018081 & 0.022004 & \cellcolor{red!20} 0.057373 \\
        $c_2$ & \cellcolor{green!20} 0.023064 & 0.032723 & \cellcolor{red!20} 0.078103 \\
        $c_3$ & \cellcolor{green!20} 0.022638 & 0.032848 & \cellcolor{red!20} 0.074036 \\
        $c_4$ & \cellcolor{green!20} 0.017847 & 0.028164 & \cellcolor{red!20} 0.058571 \\
        $c_5$ & \cellcolor{green!20} 0.012552 & 0.023663 & \cellcolor{red!20} 0.047684 \\
        $c_6$ & \cellcolor{green!20} 0.012538 & 0.023689 & \cellcolor{red!20} 0.047703 \\
        $c_7$ & \cellcolor{green!20} 0.005338 & 0.022993 & \cellcolor{red!20} 0.039377 \\
        $c_8$ & \cellcolor{green!20} 0.004655 & 0.023984 & \cellcolor{red!20} 0.039761 \\
        $p$ &\cellcolor{green!20} 0.000882 &  0.002337 & \cellcolor{red!20} 0.002799 \\
        $p_{\text{rgh}}$ & \cellcolor{green!20} 0.006197 & 0.020865 & \cellcolor{red!20} 0.023499 \\
        $\kappa$ & \cellcolor{green!20} 0.024885 & \cellcolor{red!20} 0.121615 & 0.110423 \\
        $\nu_t$ & \cellcolor{green!20} 0.020472 & 0.043081 & \cellcolor{red!20} 0.051927 \\
        $q'''$ & \cellcolor{green!20} 0.004104 & 0.023765 & \cellcolor{red!20} 0.039917 \\
        \specialrule{1.2pt}{0pt}{0pt}  
    \end{tabular}
    \caption{Comparison of the relative error $\varepsilon_2$ for each field in the state vector $\mathcal{V}$, for the different sensing strategies (out-core, mobile sensors and mobile probes). The green colouring indicates the lowest relative error for that field, whereas the red one the highest. It is clear that measurements of the flux from the solid reflector region is by far the best option in this context, with respect to use in-core mobile sensors/probes.}
    \label{tab:errors-sens-strategies}
\end{table}

For each SHRED configuration, the mobile sensor follows 1 "particle", from which the position and the value of the first group of precursor concentrations along the trajectory are collected as time-series measurements; for the mobile probe case, each configuration tracks three particles, and their positions represent the measures acting as input for SHRED. Table \ref{tab:errors-sens-strategies} shows the relative error $\varepsilon_2^\psi$ on the test set for every field in the state space $\mathcal{V}$ for the three different strategies. The best option is measuring the flux in the out-core reflector region, instead of using in-core mobile probes: the main reason to this behaviour may be due to the fact the positions of the sensors may overlap with previous time history, not allowing SHRED to properly distinguish between two different states; on the other hand, the fast flux has a monotonic exponential decay in this accidental scenario, thus avoiding the problem of bifurcation, hence showing a unique dynamical evolution which helps in the learning process of SHRED.

\begin{figure*}[tp]
    \centering
    \includegraphics[width=0.6\linewidth]{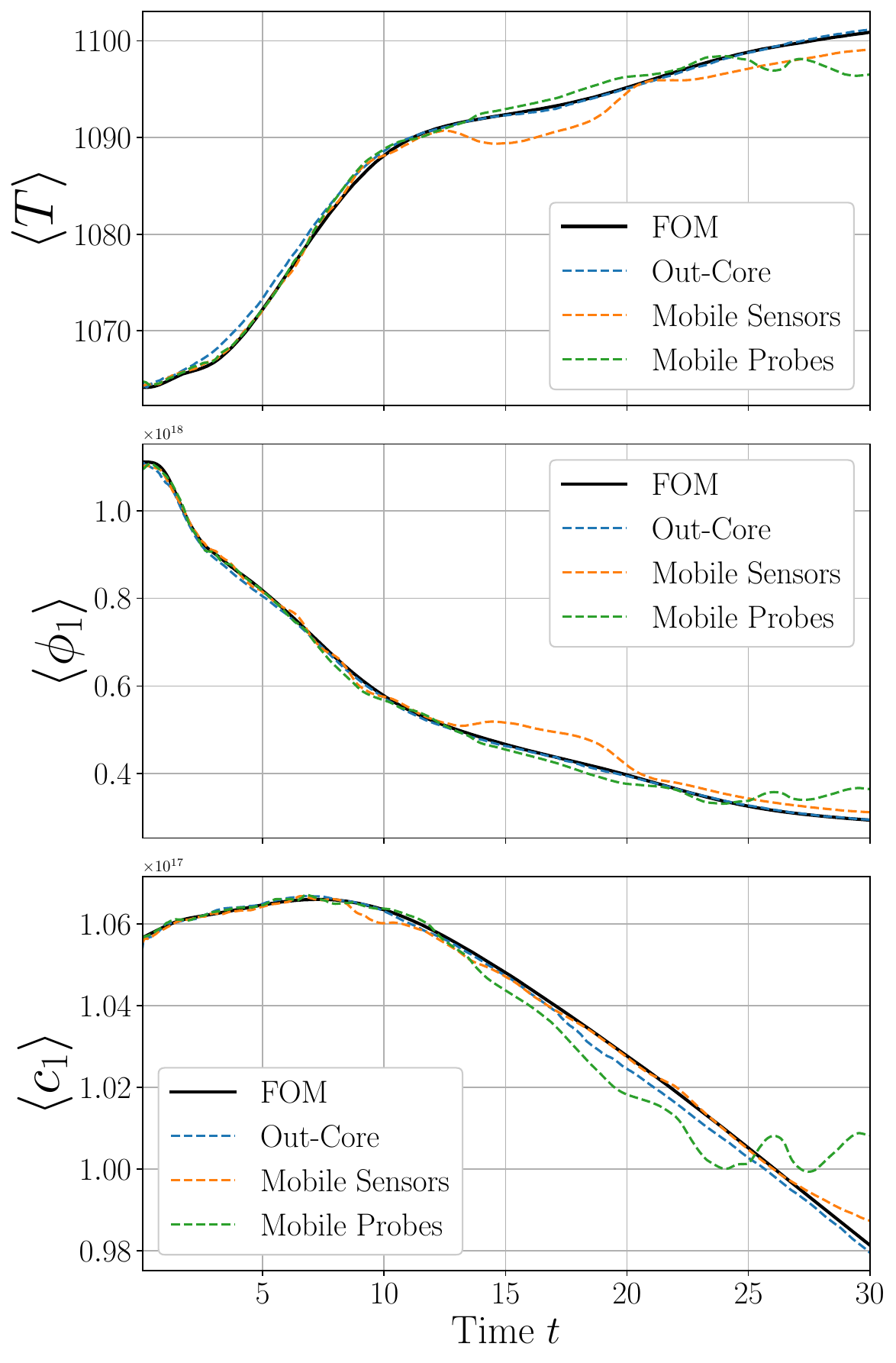}
    \caption{Comparison of the full-order and the SHRED prediction (mean) for the spatial average quantities of $T, \phi_1, c_1$ for the different sensing strategies at $\tau^\star=4.6 $ s.}
    \label{fig: averages-mobile-dynamics}
\end{figure*}

This behaviour can be further observed in Figure \ref{fig: averages-mobile-dynamics}: the average value (in space) of the temperature, fast flux, and first group of precursors are plotted against time for the test parameter $\tau^\star=4.6$ seconds. The out-core sensing strategy follows almost perfectly the dynamics of the average fields; on the other hand, the mobile sensors and probes strategies suffer a bit, especially from time $t>12$ seconds: in particular, the latter becomes inaccurate at the end of the transient due to possible non-unique behaviour, i.e., the same input can produce different outputs. Nevertheless, these sensing strategies still represent an interesting and novel option for in-core monitoring of nuclear reactors, because, supposing to be able to track the trajectory and the emission of a specific radionuclide or set of radionuclides, for example by counting the gamma ray emission from outside the primary loop, the reconstruction error is around 5-10\% as shown in Table \ref{tab:errors-sens-strategies}. Between the two mobile sensing strategies, having in the input vector for SHRED a direct measure of a quantity of interest is beneficial, especially in this "circular-system", where the particle can be in the same position at different times; the trajectory of the particle can be considered as an indirect measure of the velocity field. Moreover, as already said, particles can deposit at the boundary and get stuck, producing a constant and thus useless (for SHRED) signal.

\begin{figure*}[tp]
    \centering
        \begin{overpic}[width=0.85\linewidth]{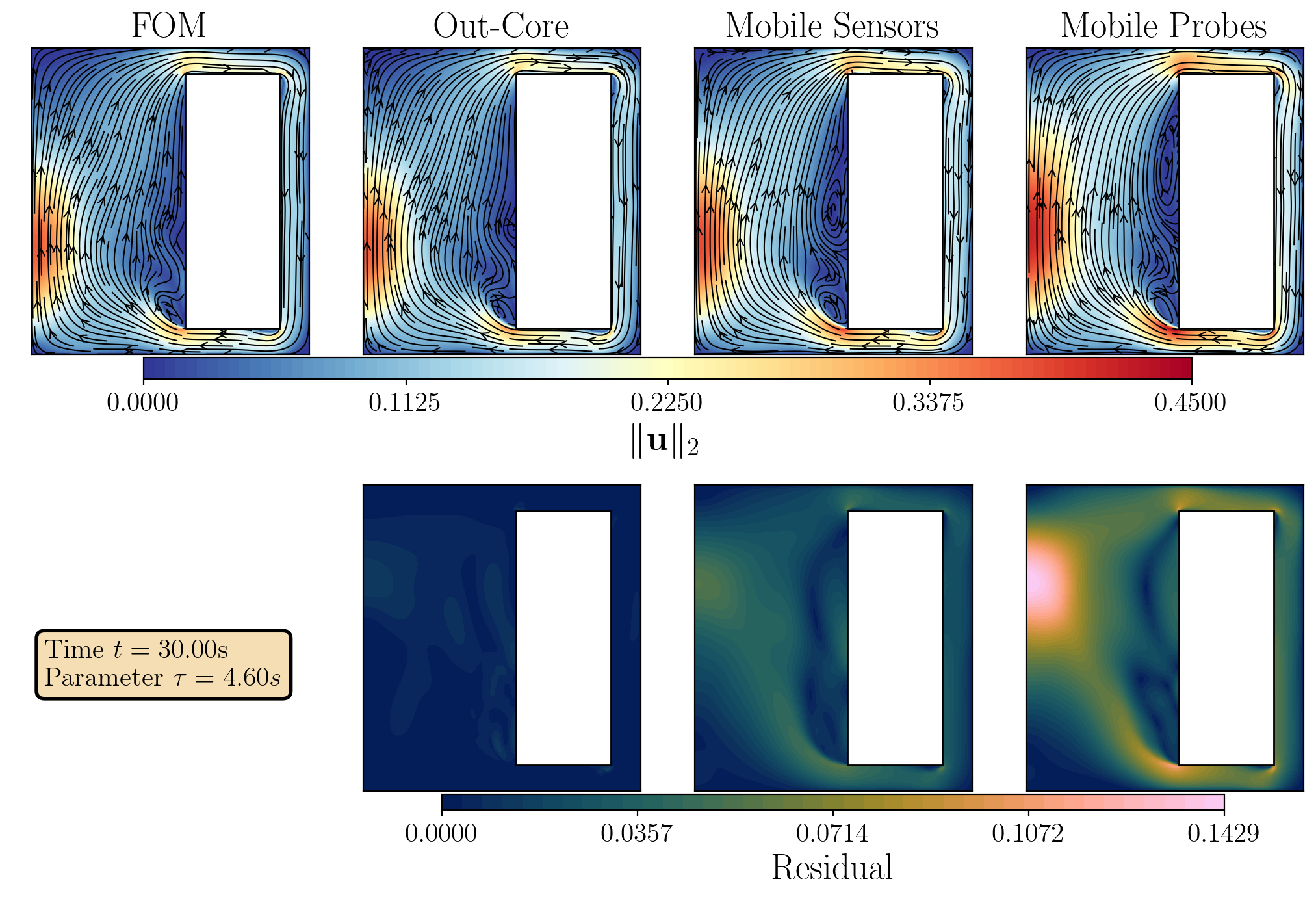}
        \put(-5,32){(a)}   
        \end{overpic}
        \begin{overpic}[width=0.85\linewidth]{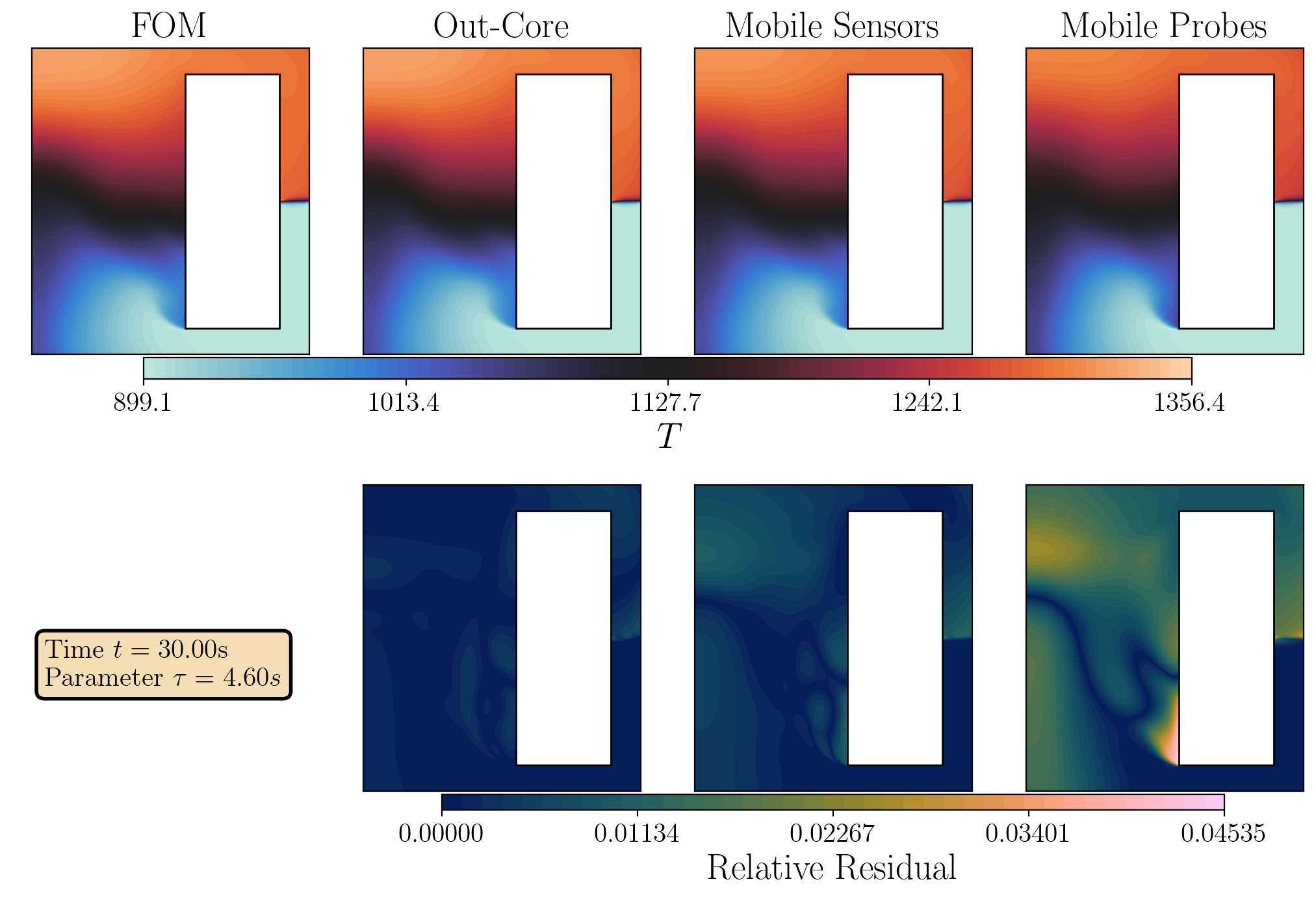}
        \put(-5,32){(b)} 
        \end{overpic}
    \caption{\rev{Contour plots at the last time step for the test parameter $\tau^*$ = 4.60 s of the velocity $\vec{u}$ (a) and the temperature $T$ (b), both unobserved. The different sensing strategies (out-core, mobile sensors and mobile probes) are compared and the prediction (mean) of the SHRED models is quite close to the ground truth, represented by the FOM: even though the mobile strategies show some more evident discrepancy, with respect to the out-core one. The last row represents the residual (absolute for (a) and relative for (b).}}
    \label{fig: mobile-contours-th}
\end{figure*}

\begin{figure*}[tp]
    \centering
        \begin{overpic}[width=0.85\linewidth]{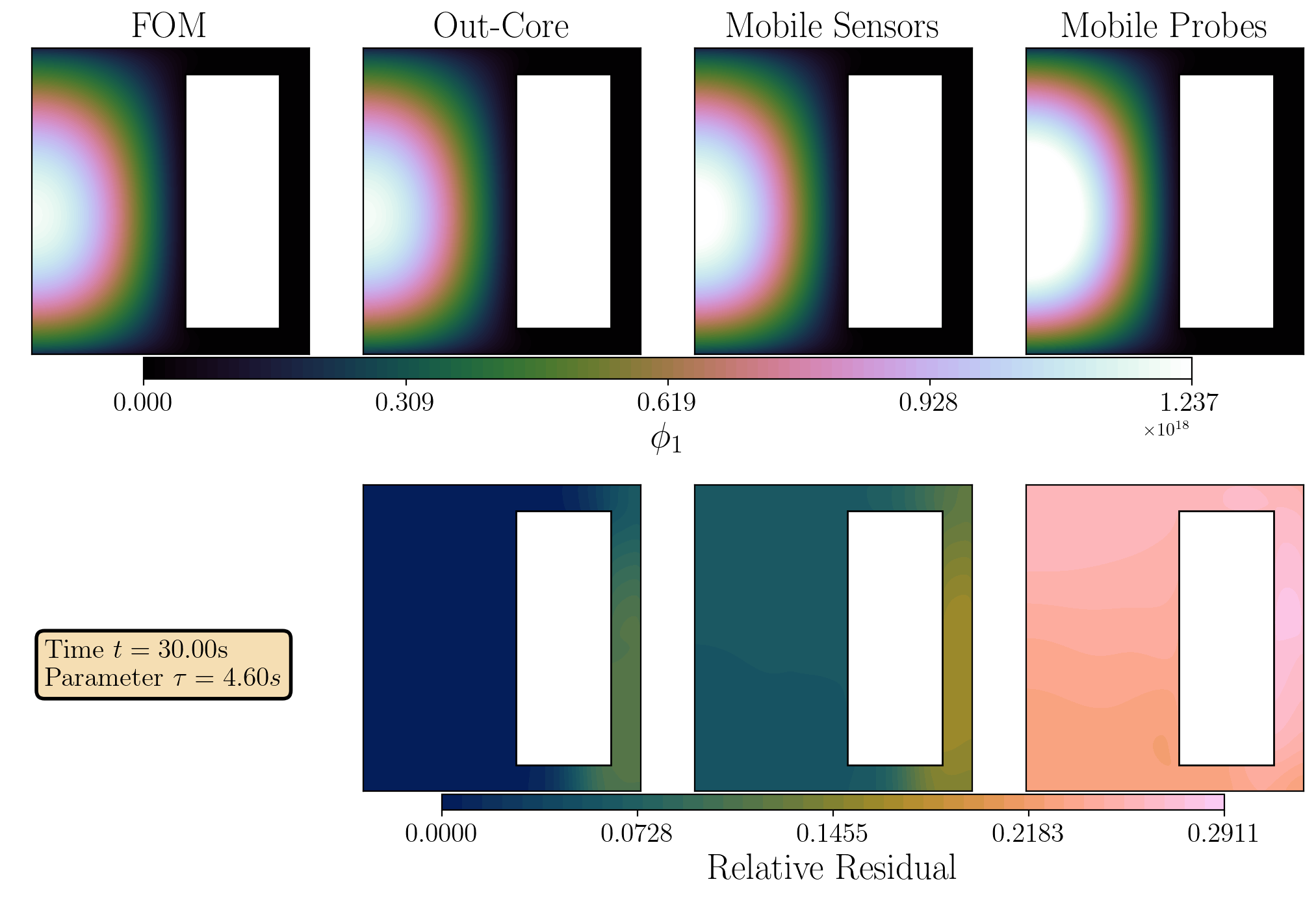}
        \put(-5,32){(a)}   
        \end{overpic}
        \begin{overpic}[width=0.85\linewidth]{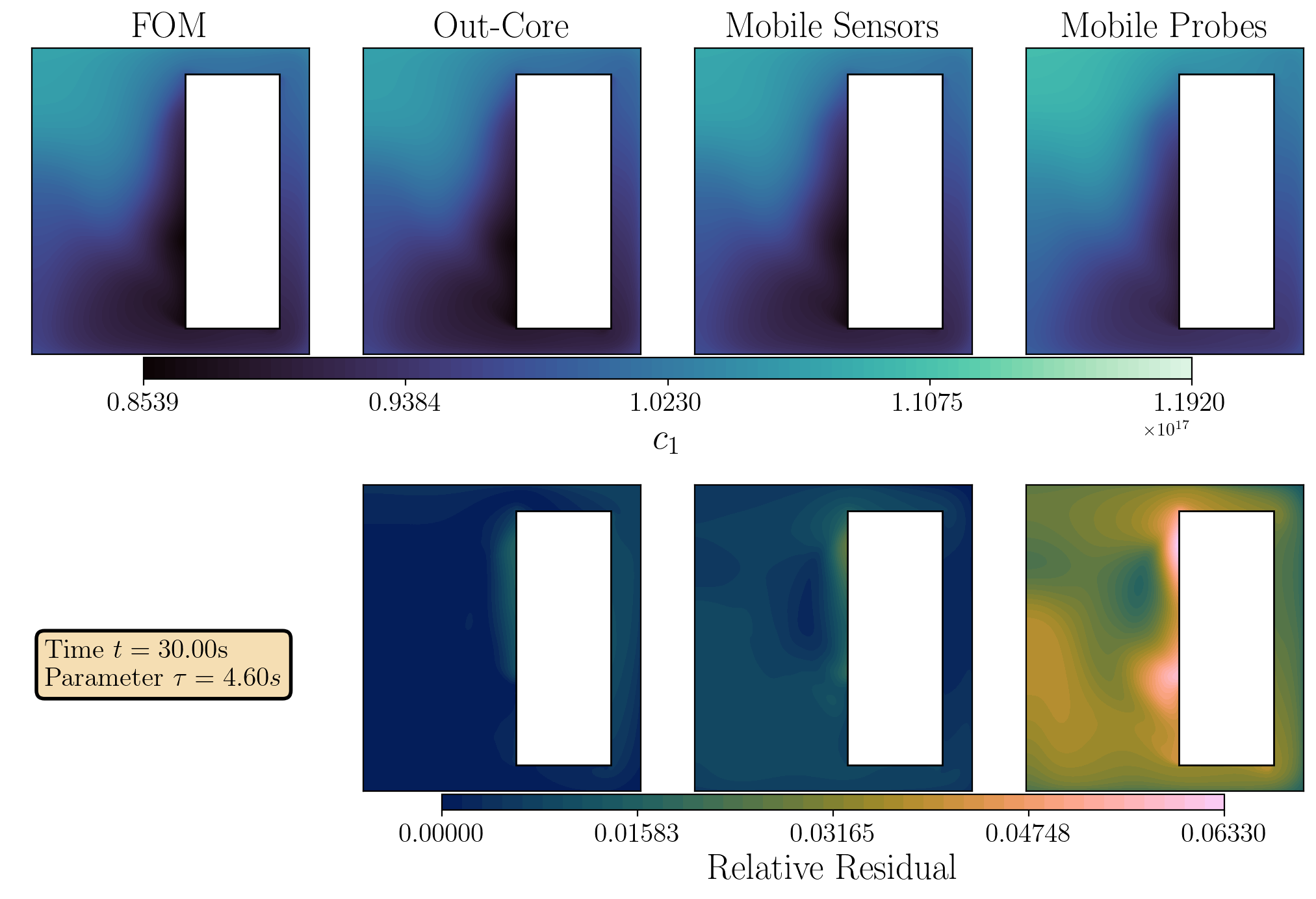}
        \put(-5,32){(b)} 
        \end{overpic}
    \caption{\rev{Contour plots at the last time step for the test parameter $\tau^*$ = 4.60 s of the fast flux $\phi_1$ (a) and first group of precursors $c_1$. The different sensing strategies (out-core, mobile sensors and mobile probes) are compared and the prediction (mean) of the SHRED models is quite close to the ground truth, represented by the FOM: even though the mobile strategies show some more evident discrepancy, with respect to the out-core one.  The last row represents the relative residual field.}}
    \label{fig: mobile-contours-neu}
\end{figure*}

In the end, the reconstructions of the velocity, temperature, and fast flux are compared to the ground-truth, i.e., the FOM, in Figure \ref{fig: mobile-contours-th} \rev{and \ref{fig: mobile-contours-neu}} at the final time for the test parameter $\tau^\star=4.6$ s. The temperature is very well reconstructed by all the strategies, showing a quite good agreement with the FOM, whereas the velocity field, \rev{the fast flux and} the precursors are not entirely accurately estimated by the mobile strategies, \rev{since it presents the highest residual among the three strategies}: in fact, for the former there are some low scale structures that are not present in the FOM and for the latter the mobile probes overestimates a bit the average value, as already highlighted in Figure \ref{fig: averages-mobile-dynamics}.

\section{Conclusions}\label{sec: concl}

This work presents an in-depth investigation of the SHallow REcurrent Decoder network architecture for the state estimation of observable and unobservable quantities of interest for a parametric accidental scenario of the Molten Salt Fast Reactor. The reactor itself is a complex engineering system that poses several challenges from both the design and process monitoring points of view, especially because of the liquid nature of the fuel, which makes conventional in-core sensing a nearly impossible task. From the methodological point of view, three different potential sensing strategies have been discussed: 1) fixed sensors in the solid out-of-core reflector region, measuring the fast flux in three random locations; 2) tracking a mobile sensor and recording its position and the concentration of the first group of precursors along the trajectory; 3) tracking three mobile probes in the liquid core by measuring only their positions over time. The SHRED architecture is used to reconstruct the whole state for these different strategies; the Unprotected Loss of Fuel Flow accidental scenario is analysed for different values of the decay constant of the pump velocity, showing how the SHRED can be naturally used for parametric problems with minimal modification of the network. The results obtained are very good and promising: in fact, there is a good agreement between the SHRED prediction and the simulation data, both in terms of latent dynamics and high-dimensional estimation, even for parameters not included in the training database. This is especially true for the out-of-core sensing, which has been shown to be superior to in-core mobile sensors and probes, even though the results are promising for each strategy. Moreover, the relatively low training time (even for parametric cases) allows for obtaining an ensemble of different models, making the prediction more robust against random noise by producing a mean value and an associated standard deviation for the estimation. 

This methodology can be used on a physical system to monitor in real-time all the quantities of interest, starting from sparse measurements of a single field. This work assumes that the model is the ground truth, and measures are taken as synthetic data polluted by noise. In the future, this hypothesis will be removed and an application to a real facility/reactor is foreseen, including a discussion on the possibility of updating the knowledge of models with measurements within the SHRED architecture. Furthermore, it is foreseen to adopt the same methodology to train a single framework able to reconstruct not a single accidental scenario, like ULOFF, but a variety of them, making the architecture able to estimate the state under different conditions.

\section*{Code and supplementary materials}  
The code and data (compressed) are available at: \href{github.com/ERMETE-Lab/NuSHRED}{github.com/ERMETE-Lab/NuSHRED}.

\section*{Acknowledgments} 
The contribution of Nathan Kutz was supported in part by the US National Science Foundation (NSF) AI Institute for Dynamical Systems (dynamicsai.org), grant 2112085.  JNK also acknowledges support from the Air Force Office of Scientific Research  (FA9550-24-1-0141)

\section*{List of Symbols}
\input{nomenclature}
\footnotesize{ \printnomenclature }

\appendix
\section{Analysis of incremental and hierarchical SVD to the MSFR dataset}\label{appendix}

The SHRED architecture has been proved to be highly efficient and generally not computationally demanding even during the training process, due to the minimal hyperparameters tuning needed (the same architecture can be adopted for different datasets \cite{williams2022data, riva2024robuststateestimationpartial, kutz_shallow_2024, shredrom}) and the possibility of working in a compressed space by projecting the snapshots onto a set of SVD modes. Regarding the latter, the limitation then lies in the size of the original dataset upon which SVD is performed: for laptop-level training, the RAM or even the available storage may not be enough. Therefore, it becomes crucial to understand how big matrices $\mathbb{X}\in\mathbb{R}^{\mathcal{N}_h\times N_p\cdot N_t}$, both in terms of number of rows (related to fine spatial discretisations, $\mathcal{N}_h>>1$) and number of columns (related to the parametric discretisation, $N_p\cdot N_t>>1$) can be optimally handled.

As already mentioned, the reduced SVD representation of $\mathbb{X}$ of rank $r$ is written as
\begin{equation}
    \mathbb{X}\simeq \mathbb{U}\Sigma\mathbb{V}^T
\end{equation}
with $\mathbb{U}\in\mathbb{R}^{\mathcal{N}_h\times r}$, $\Sigma\in\mathbb{R}^{r\times r}$ and $\mathbb{V}\in\mathbb{R}^{N_p\cdot N_t\times r}$. The randomised version of the SVD \cite{halko_finding_2010, bach_randomized_2019} is nowadays considered a very powerful algorithm able to reduce the computational times by exploiting randomisation and perform the decomposition of a partial matrix. In particular, this method uses random sampling to identify a subspace that captures most of the content of a matrix. However, this algorithm, as it has been implemented in state-of-the-art packages like scikit-learn \cite{scikit-learn}, requires storing the full-order matrix $\mathbb{X}$ in the RAM, and computational issues start arising when the matrix $\mathbb{X}$ becomes too large. 

The dataset adopted in this work falls into this category: for instance, the velocity field matrix alone would occupy about 30 GB of RAM. Therefore, to deal with this problem and thus maintain reasonably fast training times, two different variants of the SVD have been studied and compared with the randomized algorithm: an incremental SVD \cite{goos_incremental_2002}, designed to update the modes and the singular values as new data arrives by exploiting linear algebra arguments to obtain a new set of modes and singular values, and an hierarchical SVD \cite{iwen_distributed_2016}, conceived to obtain a general basis by performing sequential SVDs to favour parallel usage on GPUs. For both versions, the parametric datasets represents a perfect test case: for the former, the basis is updated as a new parametric solution arrives; for the latter, the SVDs of each parametric snapshot matrix can be performed in parallel and later another SVD can be executed on the parametric modes and singular values (see Figure 2 in \cite{iwen_distributed_2016}). 

Since the aim of this appendix consists of providing a simple but instructive analysis on how to handle big datasets coming from the numerical solution of PDEs, interested readers are invited to have a look at the cited papers for more details about the algorithms. Furthermore, it is worth mentioning the existence of pyLOM \cite{eiximeno_pylom_2025}, a Python package that can perform randomised SVD on large matrices, whose integration with SHRED will be a matter of future studies.

The comparison of the different SVD algorithms (randomized, rSVD; incremental, IncSVD; hierarchical, hSVD) will be performed in terms of computational time and relative training error compared to the starting dataset; then, the singular values, the modes and the reduced coefficients calculated by each algorithm will be analysed. For the sake of brevity, out of the 22 coupled fields in $\mathcal{V}$, only the temperature field $T$, rescaled to $[0,1]$ as in Eq. \eqref{eqn: rescaling}, has been considered; hence, $\mathbb{X}$ represents its parametric spatio-temporal behaviour during the ULOFF accident in the MSFR.

\begin{figure}[tp]
    \centering
    \includegraphics[width=1\linewidth]{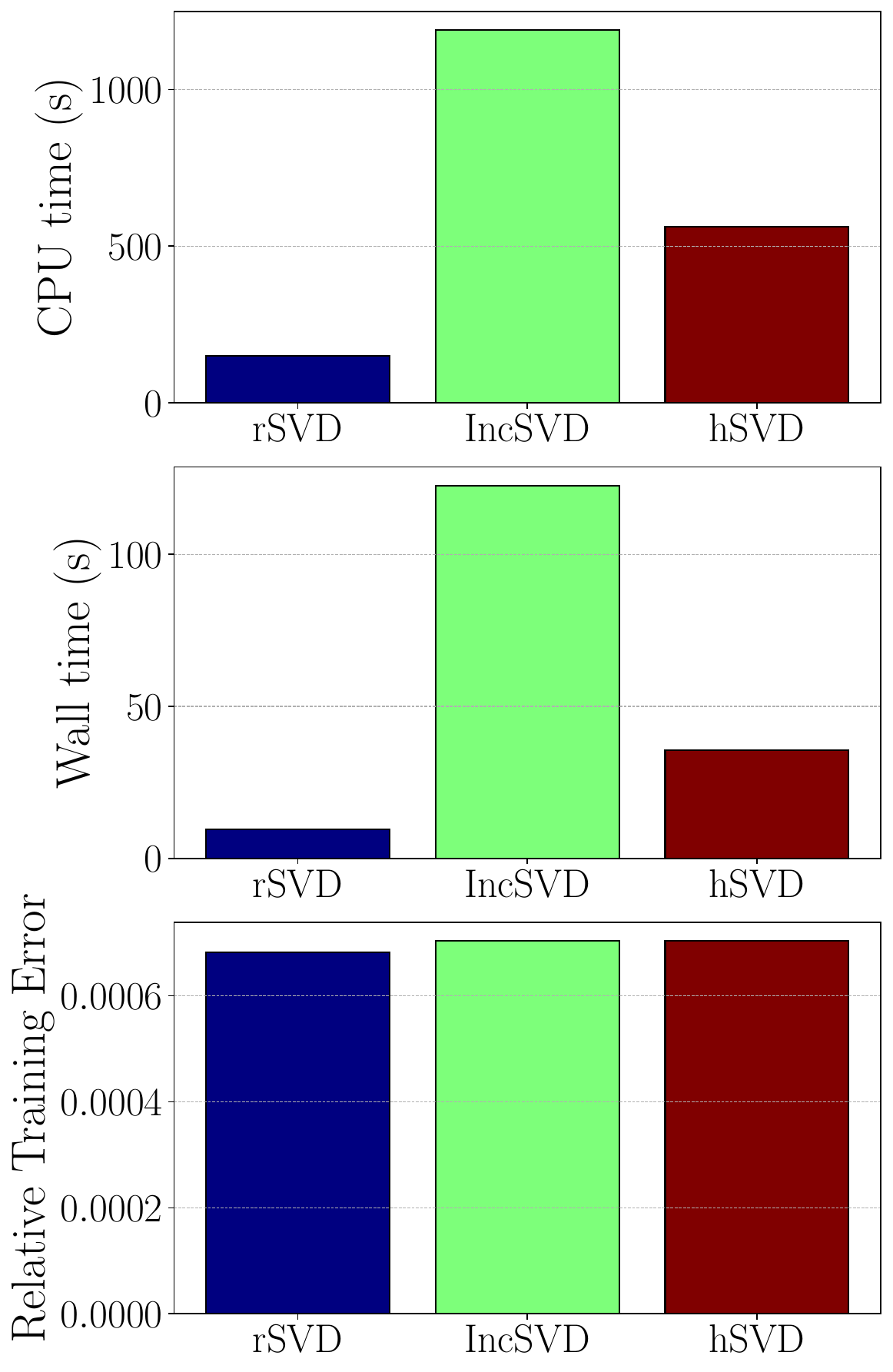}
    \caption{Comparison of the CPU time, wall time and relative training error (Euclidean norm) for the different algorithms rSVD, IncSVD and hSVD. All of them can represent the dataset with comparable error; the randomised version is cheaper from the computational point of view, followed by the hSVD and IncSVD, however it has the problem of needing to store within the RAM the entire full-order dataset matrix.}
    \label{fig: app-svd-compt-err}
\end{figure}

The matrix $\mathbb{X}$, with the temperature field, has been decomposed using the different algorithms of the SVD. Figure \ref{fig: app-svd-compt-err} shows a comparison of the computation time, in terms of CPU time and wall-clock time, and relative reconstruction error (averaged in time and parameters and measured with the energy/euclidean norm for space). Considering a maximum rank $r_{max}=25$, all the SVD versions can represent the dataset with comparable good accuracy: even more so than the computation time, this is crucial because the primary aim of the SVD is to obtain a new set of coordinates that can compress and represent the input data, thus representing a upper-bound accuracy level for the SHRED. In terms of CPU and wall time, the randomised SVD is much better than the other algorithms, meaning that if the matrix $\mathbb{X}$ can be stored in the RAM, it is recommended to use it; the incremental version \cite{goos_incremental_2002} is much more expensive than the hierarchical version since it involves a QR decomposition of a big matrix (even thought a randomised decomposition of may be more suited \cite{ni2023randomizedalgorithmqrdecompositionbased}), thus making the hierarchical version the best choice in case of RAM overflow when performing rSVD.

\begin{figure}[tp]
    \centering
    \includegraphics[width=1\linewidth]{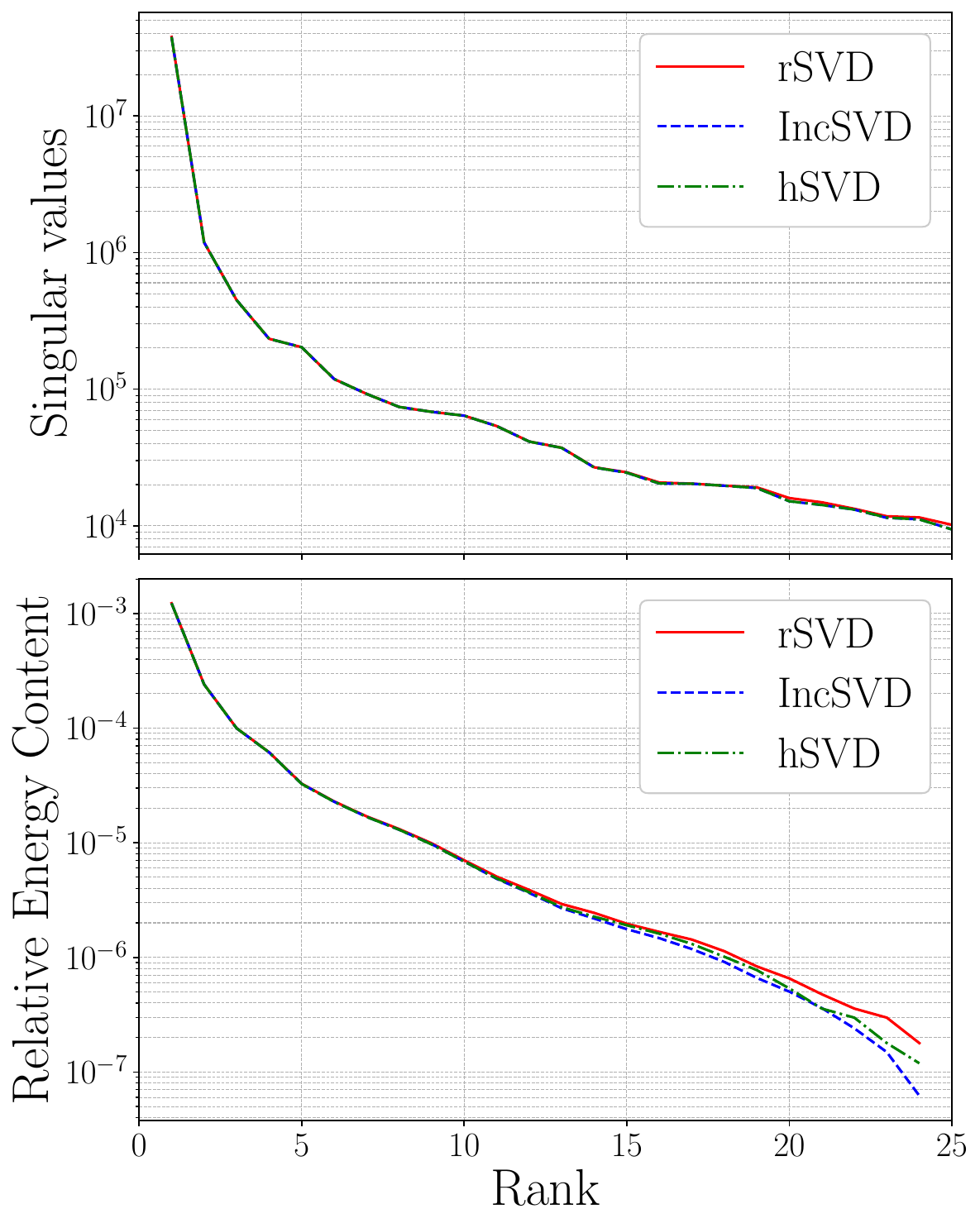}
    \caption{Decay of the singular values (top chart) and relative energy content $I(r)$ (bottom chart) as a function of the rank of the SVD. There is a very good agreement between the different algorithms with minimal discrepancy at the highest ranks.}
    \label{fig: app-svd-singvals}
\end{figure}

\begin{figure*}[tp]
    \centering
    \includegraphics[width=1\linewidth]{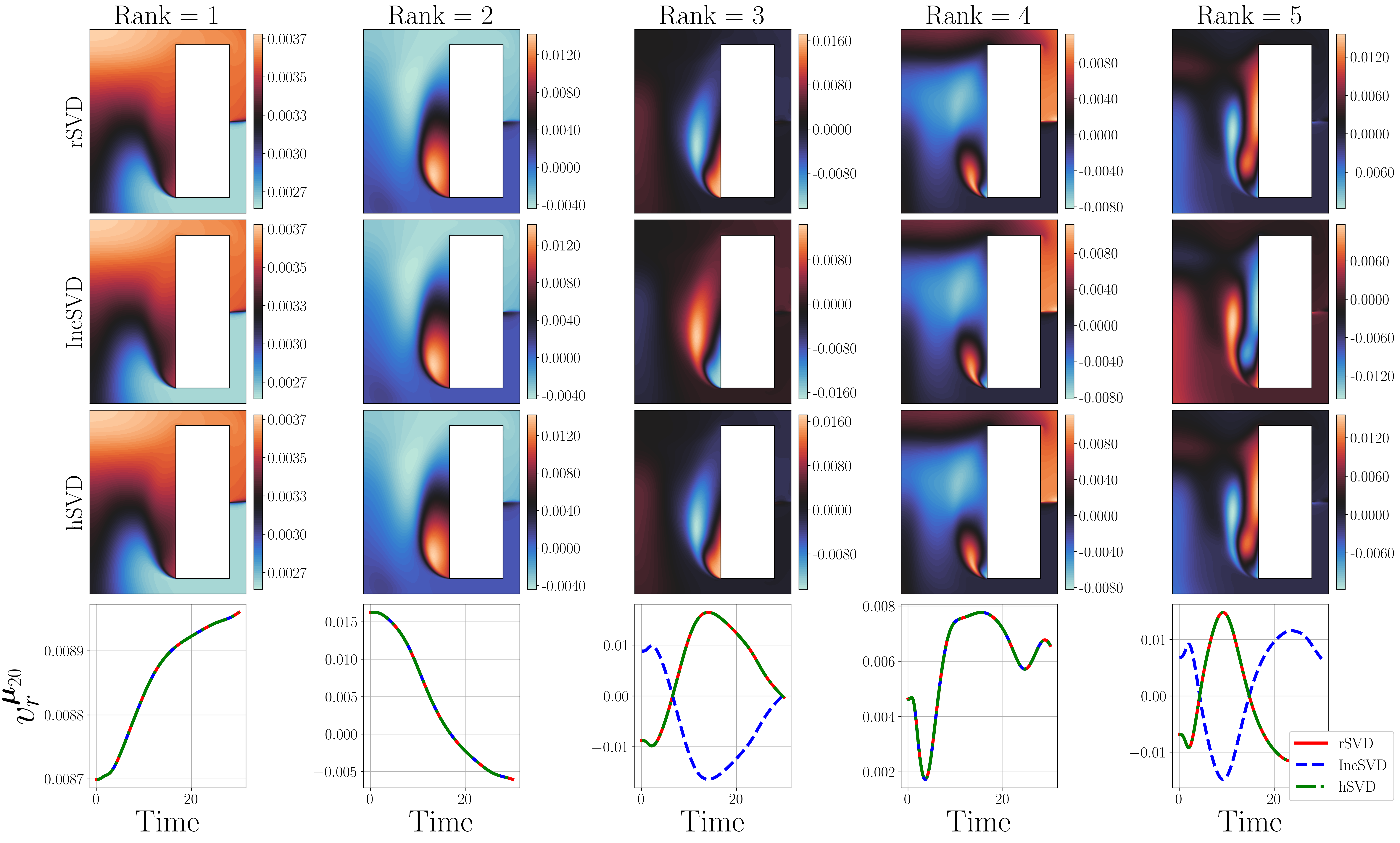}
    \caption{First 5 spatial SVD modes (rows from one to three) of the temperature field calculated by the different algorithm and the latent dynamics for a specific parameter (last row): at rank 3 and 5 the modes come with a minus sign which is reflected directly in the latent dynamics for consistency of the same original data.}
    \label{fig: app-svd-modes-dyn-comparison}
\end{figure*}

Figure \ref{fig: app-svd-singvals} shows the decay of the singular values (top chart) and the relative energy content $I(r)$ (bottom chart), defined as $I(r)=1- \frac{\sum_{k=1}^r\sigma_k^2}{\sum_{n}\sigma_n^2}$, accounting for the amount of information retained by the modes up to the chosen rank. The singular values are practically identical among the different algorithms, as well as the relative information/energy content, with very few discrepancies only for "high" ranks.

In the end, Figure \ref{fig: app-svd-modes-dyn-comparison} shows the first 5 spatial SVD modes obtained using the three algorithms. They are very similar to one another, with only two exceptions: for rank $r=3$ and $r=5$, the IncSVD predicts opposite modes. However, this minus sign is compensated by the sign of the correspondent latent dynamics (last row of the Figure), and consistency with the original dataset is retained. Thus, even in the presence of very large parametric datasets, it is still possible to operate within the compressed space and to perform laptop-level training, thanks to the alternative versions of SVD, which, at the cost of a slightly higher computation time, give comparable accuracy to the state-of-the-art randomised version.

\section{Governing Equations of the Molten Salt Fast Reactor}\label{app: msfr-gov-eqn}

The Molten Salt Fast Reactor \cite{GenIV-RoadMap} is an innovative reactor concept in which the fuel is liquid and homogeneously mixed with the molten salt thermal carrier; this configuration offers significant advantages, for example from the point of view of fuel reprocessing and closure of the fuel cycle, but also poses a lot of unique challenges in terms of monitoring and safety. Moreover, due to the strict coupling between the different physics, even more so than other nuclear reactor configurations, the mathematical modelling of such system requires suitable tools to integrate different physics together, starting from the neutronics and thermal-hydraulics \cite{aufiero2014development, CERVI2019379}.

The governing equations of the MSFR consist of the Navier-Stokes equations with the energy balance, including turbulence modelling with RANS (Realizable $\kappa-\epsilon$ model) and the Boussinesq approximation for density variations, the multi-group neutron diffusion and the precursors advection-diffusion: for completeness, this section reports the governing equations and briefly discusses their coupling. \rev{The equations that will be presented below are related to the transient problem, the initial condition is always the critical state from which the accident starts. In order to obtain the steady initial condition a $k$-eigenvalue calculation has been previously before. Interested readers can refer to \cite{aufiero2014development, CERVI2019379}.}

The Navier-Stokes equations for the MSFR reads \cite{versteeg2007introduction}:
\begin{subequations}
\begin{align}
\nabla\cdot \vec{u} =& 0 \label{eq: mass} \\
\frac{\partial\vec{u}}{\partial t} + (\vec{u}\cdot \nabla)\vec{u} =& +\nabla\cdot [\nu_{\text{eff}}(\nabla\vec{u} + (\nabla\vec{u})^{T})] \nonumber \\
&- \frac{1}{\rho} \nabla p + [1 - \beta_{T}(T - T_{0})] \vec{g}  \label{eq: velocity} \\
\frac{\partial T}{\partial t} + \nabla \cdot \left(\vec{u} T\right)  =& \nabla\cdot (\alpha_{\text{eff}}\nabla T) + \frac{q'''}{\rho c_{p}} \label{eq: energy}
\end{align}
\end{subequations}
given $\vec{u}$ as the velocity, $\rho$ as the density, $p$ as the pressure, $\nu_{\text{eff}}$ as the effective kinematic viscosity (accounting for turbulence modelling, \rev{i.e. $\nu+\nu_T$)}, $T$ as the temperature, $\beta_T$ as the thermal expansion, $\vec{g}$ as the gravity acceleration, $\alpha_{\text{eff}}$ as the \rev{effective} thermal diffusivity \rev{(accounting for turbulence modelling, i.e. $\frac{\nu}{Pr}+\frac{\nu_T}{Pr_T}$, with $Pr$ stands for the Prandtl number)}, $q'''$ as the power density, which includes both neutron fission and delayed heat sources, and $c_p$ as the specific heat capacity. \rev{The PDE system is closed by suitable boundary conditions, in particular no slip condition for the velocity and adiabatic for the temperature on the outer boundary}, rigorously stated in \ref{app: msfr-bc-ic}\rev{.} This set of equations shows the strong coupling between the two different physics; in particular, the source term of the energy equation directly depends on the neutron fluxes, i.e.
\begin{equation}
    q''' = (1-\beta_{h})\sum_{g}\bar{E}_{f,g}\Sigma_{f,g}\phi_{g} + \sum_{i}\lambda_{h,i}d_{i}
    \label{eqn: volumetric-heat-source}
\end{equation}
with $\phi_g$ being the neutron flux of energy group $g$, $\beta_h$ being the fraction of delayed heat source, $\bar{E}_{f,g}$ being the average energy released by fission in group $g$, $\Sigma_{f,g}$ being the fission cross section \cite{DuderstadtHamilton} of group $g$, $\lambda_{h,i}$ being the  decay constant of delayed decay group $d_i$.

Neutronics is governed by the Multi-Group Diffusion Equation \cite{DuderstadtHamilton, aufiero2014development}:
\begin{equation}	
    \frac{1}{v_{g}} \frac{\partial\phi_{g}}{\partial t} - \nabla\cdot (D_{n,g} \nabla \phi_{g}) + \Sigma_{r,g} \phi_{g} = S_{g} 
\label{eqn: diffusion-mg}
\end{equation}
with $D_{n,g}$ being the neutron diffusion coefficient of group $g$ and $\Sigma_{r,g}$ being the removal cross-section accounting for all the reactions which decrease the number of neutrons of group $g$, i.e. absorption $\Sigma_{a,g}$ and infra-group scattering $\Sigma_{s,g \to g'}$:
\begin{equation}
    \Sigma_{r,g}^0 =  \left(\Sigma_{a,g}^0+ \sum_{ g' \neq g }\Sigma_{s,g \to g'}^0 \right)
\end{equation}
where the superscript $0$ refers to the values at the reference temperature. In fact, the cross section are strongly affected by the neutron energy due to the Doppler broadening effect and the variation in density, i.e. 
\begin{equation}
    \Sigma_{r,g} = \left( \Sigma_{r,g}^{0} + A_{r,g}^{0}\ln\frac{T}{T_{0}^{\Sigma}}\right)\cdot \frac{1-\beta_{T}(T - T_{0})}{1 - \beta_{T}(T_{0}^{\Sigma} - T_{0})}
    \label{eqn: xs-mg}
\end{equation}
\rev{with $A_{r,g}^{0}$ as a fitting coefficient for the log-law and $T_{0}^{\Sigma}$ is the reference temperature at which the reference group cross section $\Sigma_{r,g}^{0}$ is calculated.} In the end, the neutron diffusion equations have a source term $S_g$ accounting for the production of neutrons in the group energy $g$ due to prompt fission, decay of precursors and infra-group scattering:
\begin{subequations}
\begin{align}
    S_{g} &= (1-\beta) \chi_{p,g}\,S_{p,g} +\chi_{d,g}\,S_{d} + S_{s,g}\\
      S_{p,g} &= \sum_{g'}\bar{\nu}_{g'}\Sigma_{f,g'}\phi_{g'} \\
      S_{d}   &=\sum_{k}\lambda_{k}c_{k} \\
      S_{s,g} &= \sum_{g' \neq g}\Sigma_{s,g' \to g} \phi_{g'}
\end{align}
\end{subequations}
with $\chi_{p,g}$ and $\chi_{d,g}$ as the fission spectrum of prompt and delayed neutrons and $\bar{\nu}_{g'}$ as the average number of neutrons produced in group $g'$. \rev{The diffusion equations for the neutron flux is closed by albedo boundary conditions} (see \ref{app: msfr-bc-ic})\rev{.}

When dealing with transient scenarios, the production of neutrons is split into prompt (due to fission) and delayed (due to decay): in the nuclear reactors world, this latter contribution is described by the group precursors $c_k$ which are governed by an advection-diffusion equation in the salt \rev{(closed by homogenous Neumann condition at the outer boundary}, see \ref{app: msfr-bc-ic}\rev{)}:
\begin{equation}
\begin{split}
\frac{\partial c_{k}}{\partial t} + \nabla\cdot (\vec{u} c_{k}) =& +\nabla\cdot (D_{\text{eff}} \nabla c _{k}) - \lambda_{k} c _{k} \\
&+ \beta_{d,k}\sum_{g} \bar{\nu}_{g} \Sigma_{f,g}\phi_{g}
\end{split}
\label{eqn: precursors}
\end{equation}
with $D_{\text{eff}}$ the effective diffusion coefficient \rev{(accounting for turbulence modelling, i.e. $\frac{\nu}{Sc}+\frac{\nu_T}{Sc_T}$, with $Sc$ stands for the Schmidt number)} and $\lambda_k$ the decay constant of delayed neutron precursor group $k$. In the end, the precursors contribute also to the energy equation \eqref{eq: energy}, where they are grouped together into the decay heat group $d_i$
\begin{equation}
\begin{split}
    \frac{\partial d_{i}}{\partial t} + \nabla\cdot(\vec{u}d_{i}) =& + \nabla\cdot(D_{\text{eff}}\nabla d_{i}) - \lambda_{h,i}d_{i} \\
    &+ \beta_{h,l}\sum_{g}\bar{E}_{f,g}\Sigma_{f,g}\phi_{g}
\end{split}
\label{eqn: decay-heat-prec}
\end{equation}

\subsection{Boundary and initial conditions}\label{app: msfr-bc-ic}

The primary circuit is modelled as a single, computational domain $\Omega$, as depicted in Figure \ref{fig: evol-geom}: no interface is introduced between the active core and the outer loop. The fuel salt, together with the delayed neutron precursors, circulates continuously through the core, the heat exchanger and the pump; no internal boundary condition is required between them. Consequently, the action of the pump and of the heat exchanger is prescribed as a volumetric source terms localised in the corresponding cell zones. The domain is partitioned into the liquid fuel region $\Omega_f$ (light blue in Figure \ref{fig: evol-geom}) and the solid reflector $\Omega_r$ (dark blue): the neutronic fields and the temperature (through heat conduction) are solved over the whole $\Omega=\Omega_f\cup\Omega_r$, whereas the flow fields (velocity and turbulent quantities) are confined to the fluid region $\Omega_f$, the velocity being constrained to zero in the solid reflector $\Omega_r$. The only physical boundaries of the problem are therefore the external walls of the domain, collectively denoted by $\partial\Omega_w$; these correspond, in Figure \ref{fig: evol-geom}, to the outer contour of the reflector, the inner contour surrounding the non-simulated fertile blanket, and the top and bottom walls of the loop. The left lateral face $\partial \Omega_s$ of the $5^\circ$ wedge is treated as symmetry plane. In the following, $\vec{n}$ denotes the outward unit normal to $\partial\Omega = \partial \Omega_w \cup\partial \Omega_s$.

\emph{Thermal-hydraulics.} The Navier--Stokes system, Eqs.~\eqref{eq: mass}--\eqref{eq: energy}, is closed on $\partial\Omega_w$ by a no-slip condition for the velocity, an adiabatic (zero wall-normal gradient) condition for the temperature, a zero wall-normal gradient for the pressure (consistent with the impermeable walls of the closed loop) and standard low-Reynolds wall functions for the turbulent quantities \cite{aufiero2014development}:
\begin{subequations}
\begin{align}
\vec{u} &= \vec{0} && \text{on } \partial\Omega_w, \\
\vec{n}\cdot\nabla T &= 0 && \text{on } \partial\Omega_w, \\
\vec{n}\cdot\nabla p &= 0 && \text{on } \partial\Omega_w, \\
\kappa,\ \varepsilon,\ \nu_t &: \ \text{wall functions} && \text{on } \partial\Omega_w .
\end{align}
\label{eqn: bc-th}
\end{subequations}

\emph{Neutronics.} Each of the six group fluxes, Eq.~\eqref{eqn: diffusion-mg}, is closed on $\partial\Omega$ by an albedo boundary condition, which prescribes the ratio between the incoming and outgoing partial currents at the wall through the group albedo $\gamma_g\in[0,1]$:
\begin{equation}
-D_{n,g}\,\vec{n}\cdot\nabla\phi_g = \frac{1-\gamma_g}{2\,(1+\gamma_g)}\,\phi_g \qquad \text{on }\partial\Omega_w,\quad g=1,\dots,6,
\label{eqn: bc-flux}
\end{equation}
where the limit $\gamma_g\to 1$ recovers a perfectly reflective wall (zero net current) and $\gamma_g\to 0$ a vacuum (Marshak) boundary. In the present model a vacuum-like albedo is imposed at the outer boundary of the reflector, so that the net leakage current is proportional to the local flux.

\emph{Delayed neutron and decay-heat precursors.} The transported precursor fields are closed by a homogeneous Neumann condition on the whole boundary: no diffusive flux of precursors leaves the domain, as they are only transported around the closed loop by advection and depleted by radioactive decay,
\begin{equation}
\vec{n}\cdot\nabla c_k = 0, \quad \vec{n}\cdot\nabla d_i = 0 \quad \text{on }\partial\Omega_w, \quad k=1,\dots,8,\ i=1,\dots,3 .
\label{eqn: bc-prec}
\end{equation}

\emph{Initial conditions.} Every transient simulation starts from the steady critical configuration of the reactor at nominal (full-power) operating conditions. This state is obtained from a $k$-eigenvalue calculation of the coupled steady problem, in which the fission neutron production is divided by the effective multiplication factor $k_{\text{eff}}$ so that a stationary solution of the neutron balance exists, and the precursors are advected by the nominal velocity field $\vec{u}$, obtained by solving a steady version of the Navier--Stokes equations. The same boundary conditions \eqref{eqn: bc-flux}--\eqref{eqn: bc-prec} apply. This is the well-established formulation adopted for the MSFR \cite{aufiero2014development, CERVI2019379}, and it guarantees that the transient starts exactly from a critical, self-consistent multiphysics state.

\subsection{Adopted simulation parameters}\label{app: msfr-parameters}

To ensure full reproducibility of the results, the parameters adopted in the OpenFOAM multi-physics simulations are collected in this section. All the quantities are reported in SI units and refer to the symbols introduced in the governing equations above. The values are given for the two material regions of the model, i.e. the circulating liquid fuel salt (the $^7$LiF--$^{232}$ThF$_4$ eutectic) and the solid Hastelloy reflector; the six energy-group macroscopic cross sections, the associated Doppler feedback coefficients and the kinetic data have been generated as described in \cite{aufiero2014development, CERVI2019379}. The initial condition of each transient is the critical steady state obtained from a $k$-eigenvalue calculation.

\newcommand{\ee}[2]{#1\cdot 10^{#2}}

\begin{table*}[tp]
\centering
\scriptsize
\begin{tabular}{llccc}
\toprule
Parameter & Symbol & Fuel salt & Reflector & Unit \\
\midrule
Density                       & $\rho$          & $4306.71$        & $10000$          & kg\,m$^{-3}$ \\
Specific heat capacity        & $c_p$           & $1593.94$        & $600$            & J\,kg$^{-1}$\,K$^{-1}$ \\
Laminar kinematic viscosity   & $\nu$           & $\ee{5.89}{-6}$  & --               & m$^2$\,s$^{-1}$ \\
Thermal expansion coefficient & $\beta_T$       & $\ee{1.912}{-4}$ & $\ee{1.38}{-5}$  & K$^{-1}$ \\
Laminar Prandtl number        & $Pr$            & $23.78$          & $1.048$          & -- \\
Turbulent Prandtl number      & $Pr_t$          & $0.85$           & $\ee{1.0}{8}$    & -- \\
Laminar Schmidt number        & $Sc$            & $20$             & --               & -- \\
Turbulent Schmidt number      & $Sc_t$          & $0.85$           & --               & -- \\
Boussinesq reference temp.    & $T_0$           & $973$            & --               & K \\
Doppler reference temp.       & $T_0^\Sigma$    & $900$            & --               & K \\
Gravity acceleration          & $g$             & $9.81$           & --               & m\,s$^{-2}$ \\
\midrule
Total (full-core) thermal power & $P$           & \multicolumn{2}{c}{$\ee{3.0}{9}$}   & W \\
Pump momentum source (nominal)  & --            & \multicolumn{2}{c}{$-26.6$}         & m\,s$^{-2}$ \\
Heat-exchanger exchange coeff.  & $h$           & \multicolumn{2}{c}{$\ee{2.5}{6}$}   & W\,m$^{-3}$\,K$^{-1}$ \\
Heat-exchanger secondary temp.  & $T_{ext}$     & \multicolumn{2}{c}{$900$}           & K \\
\bottomrule
\end{tabular}
\caption{Thermo-physical and operational parameters adopted in the simulations. The full core power refers to the whole reactor, while the modelled domain is a $5^\circ/360^\circ$ axisymmetric wedge. During the ULOFF transient the pump momentum source is decreased exponentially as $\sim e^{-t/\tau}$, with $\tau\in[1,10]$\,s ($N_p=21$ values).}
\label{tab:params-thermo}
\end{table*}

\begin{table*}[tp]
\centering
\footnotesize
\begin{adjustbox}{max width=\textwidth}
\begin{tabular}{ccccccccc}
\toprule
$g$ & $D_{n,g}$ & $\Sigma_{a,g}^0$ & $\Sigma_{f,g}^0$ & $\bar{\nu}_g$ & $\bar{E}_{f,g}$ & $1/v_g$ & $\chi_{p,g}$ & $\chi_{d,g}$ \\
\midrule
$1$ & $\ee{2.315}{-2}$ & $\ee{6.211}{-1}$ & $\ee{4.446}{-1}$ & $3.076$ & $\ee{3.304}{-11}$ & $\ee{3.981}{-8}$ & $\ee{3.607}{-1}$ & $\ee{4.645}{-3}$ \\
$2$ & $\ee{1.540}{-2}$ & $\ee{3.258}{-1}$ & $\ee{2.517}{-1}$ & $2.863$ & $\ee{3.315}{-11}$ & $\ee{7.520}{-8}$ & $\ee{5.172}{-1}$ & $\ee{3.939}{-1}$ \\
$3$ & $\ee{9.760}{-3}$ & $\ee{3.952}{-1}$ & $\ee{1.805}{-1}$ & $2.708$ & $\ee{3.299}{-11}$ & $\ee{2.688}{-7}$ & $\ee{1.206}{-1}$ & $\ee{5.754}{-1}$ \\
$4$ & $\ee{1.180}{-2}$ & $\ee{7.724}{-1}$ & $\ee{2.618}{-1}$ & $2.672$ & $\ee{3.294}{-11}$ & $\ee{6.620}{-7}$ & $\ee{1.369}{-3}$ & $\ee{2.307}{-2}$ \\
$5$ & $\ee{1.103}{-2}$ & $1.547$          & $\ee{5.200}{-1}$ & $2.667$ & $\ee{3.293}{-11}$ & $\ee{1.495}{-6}$ & $\ee{1.515}{-4}$ & $\ee{2.674}{-3}$ \\
$6$ & $\ee{1.026}{-2}$ & $3.879$          & $1.395$          & $2.672$ & $\ee{3.295}{-11}$ & $\ee{3.658}{-6}$ & $\ee{7.457}{-6}$ & $\ee{3.163}{-4}$ \\
\bottomrule
\end{tabular}
\end{adjustbox}
\caption{Six energy-group macroscopic cross sections and kinetic data of the fuel salt at the reference temperature $T_0^\Sigma=900$\,K. Cross sections and $1/v_g$ are in m$^{-1}$ and s\,m$^{-1}$, $D_{n,g}$ in m, $\bar{E}_{f,g}$ in J; $\bar\nu_g$, $\chi_{p,g}$ and $\chi_{d,g}$ are dimensionless.}
\label{tab:params-neutronics-fuel}
\end{table*}

\begin{table*}[tp]
\centering
\footnotesize
\begin{adjustbox}{max width=\textwidth}
\begin{tabular}{lcccccc}
\toprule
$\Sigma_{s,g\to g'}^0$ & $g'=1$ & $g'=2$ & $g'=3$ & $g'=4$ & $g'=5$ & $g'=6$ \\
\midrule
$g=1$ & $\ee{1.355}{1}$ & $5.805$          & $\ee{6.157}{-1}$ & $\ee{4.173}{-3}$ & $\ee{4.294}{-4}$ & $\ee{3.618}{-5}$ \\
$g=2$ & $0$             & $\ee{2.204}{1}$  & $3.766$          & $\ee{6.430}{-3}$ & $\ee{5.358}{-4}$ & $\ee{2.385}{-5}$ \\
$g=3$ & $0$             & $0$              & $\ee{3.504}{1}$  & $1.412$          & $\ee{2.366}{-3}$ & $\ee{1.490}{-4}$ \\
$g=4$ & $0$             & $0$              & $0$              & $\ee{2.719}{1}$  & $1.275$          & $\ee{2.403}{-4}$ \\
$g=5$ & $0$             & $0$              & $0$              & $0$              & $\ee{2.902}{1}$  & $\ee{5.693}{-1}$ \\
$g=6$ & $0$             & $0$              & $0$              & $0$              & $0$              & $\ee{2.955}{1}$ \\
\bottomrule
\end{tabular}
\end{adjustbox}
\caption{Group-to-group scattering cross section matrix $\Sigma_{s,g\to g'}^0$ of the fuel salt at the reference temperature (in m$^{-1}$).}
\label{tab:params-scatter-fuel}
\end{table*}

\begin{table*}[tp]
\centering
\footnotesize
\begin{tabular}{cccc}
\toprule
$g$ & $A_{D,g}^0$ & $A_{a,g}^0$ & $A_{f,g}^0$ \\
\midrule
$1$ & $\ee{-1.182}{-5}$ & $\ee{-7.786}{-4}$ & $\ee{-8.343}{-5}$ \\
$2$ & $\ee{3.824}{-6}$  & $\ee{-2.086}{-5}$ & $\ee{-2.781}{-5}$ \\
$3$ & $\ee{-3.476}{-6}$ & $\ee{2.781}{-4}$  & $\ee{6.952}{-6}$ \\
$4$ & $\ee{-3.928}{-5}$ & $\ee{4.665}{-3}$  & $\ee{8.343}{-5}$ \\
$5$ & $\ee{-1.390}{-4}$ & $\ee{4.383}{-2}$  & $\ee{-7.439}{-4}$ \\
$6$ & $\ee{-2.346}{-4}$ & $\ee{2.365}{-1}$  & $\ee{-1.046}{-2}$ \\
\bottomrule
\end{tabular}
\caption{Doppler feedback coefficients (fitting coefficients $A_{r,g}^0$ of the logarithmic law of Eq.~\eqref{eqn: xs-mg}) for the diffusion, absorption and fission cross sections of the fuel salt. Diffusion coefficients are in m, absorption/fission coefficients in m$^{-1}$.}
\label{tab:params-doppler-fuel}
\end{table*}

\begin{table*}[tp]
\centering
\footnotesize
\begin{adjustbox}{max width=\textwidth}
\begin{tabular}{lcccccc}
\toprule
$A_{s,g\to g'}^0$ & $g'=1$ & $g'=2$ & $g'=3$ & $g'=4$ & $g'=5$ & $g'=6$ \\
\midrule
$g=1$ & $\ee{-1.008}{-2}$ & $\ee{1.046}{-2}$  & $\ee{2.736}{-3}$  & $\ee{-9.611}{-5}$ & $\ee{3.701}{-5}$  & $\ee{-1.416}{-5}$ \\
$g=2$ & $0$               & $\ee{-2.086}{-3}$ & $\ee{1.529}{-3}$  & $\ee{-1.317}{-4}$ & $\ee{2.341}{-5}$  & $\ee{-3.173}{-6}$ \\
$g=3$ & $0$               & $0$               & $\ee{9.733}{-3}$  & $\ee{2.781}{-4}$  & $\ee{-3.163}{-6}$ & $\ee{1.354}{-5}$ \\
$g=4$ & $0$               & $0$               & $0$               & $\ee{8.864}{-2}$  & $\ee{-1.842}{-3}$ & $\ee{-3.230}{-5}$ \\
$g=5$ & $0$               & $0$               & $0$               & $0$               & $\ee{3.559}{-1}$  & $\ee{-1.585}{-2}$ \\
$g=6$ & $0$               & $0$               & $0$               & $0$               & $0$               & $\ee{5.242}{-1}$ \\
\bottomrule
\end{tabular}
\end{adjustbox}
\caption{Doppler feedback coefficients $A_{s,g\to g'}^0$ of the scattering matrix of the fuel salt (in m$^{-1}$).}
\label{tab:params-doppler-scatter-fuel}
\end{table*}

\begin{table*}[tp]
\centering
\footnotesize
\begin{tabular}{cccc}
\toprule
$g$ & $D_{n,g}$ & $\Sigma_{a,g}^0$ & $1/v_g$ \\
\midrule
$1$ & $\ee{1.660}{-2}$ & $1.290$          & $\ee{4.030}{-8}$ \\
$2$ & $\ee{1.500}{-2}$ & $\ee{1.280}{-1}$ & $\ee{8.160}{-8}$ \\
$3$ & $\ee{7.210}{-3}$ & $\ee{2.400}{-1}$ & $\ee{2.370}{-7}$ \\
$4$ & $\ee{2.290}{-3}$ & $\ee{6.660}{-1}$ & $\ee{7.170}{-7}$ \\
$5$ & $\ee{2.240}{-3}$ & $1.450$          & $\ee{1.650}{-6}$ \\
$6$ & $\ee{2.300}{-3}$ & $3.870$          & $\ee{4.330}{-6}$ \\
\bottomrule
\end{tabular}

\vspace{0.7em}

\begin{adjustbox}{max width=\textwidth}
\begin{tabular}{lcccccc}
\toprule
$\Sigma_{s,g\to g'}^0$ & $g'=1$ & $g'=2$ & $g'=3$ & $g'=4$ & $g'=5$ & $g'=6$ \\
\midrule
$g=1$ & $9.230$ & $7.710$          & $1.760$          & $\ee{1.260}{-2}$ & $\ee{7.950}{-4}$ & $0$ \\
$g=2$ & $0$     & $\ee{2.030}{1}$  & $1.830$          & $\ee{8.450}{-3}$ & $\ee{7.800}{-4}$ & $\ee{5.570}{-5}$ \\
$g=3$ & $0$     & $0$              & $\ee{4.520}{1}$  & $\ee{7.680}{-1}$ & $\ee{1.400}{-3}$ & $\ee{1.710}{-5}$ \\
$g=4$ & $0$     & $0$              & $0$              & $\ee{1.420}{2}$  & $3.070$          & $0$ \\
$g=5$ & $0$     & $0$              & $0$              & $0$              & $\ee{1.460}{2}$  & $1.550$ \\
$g=6$ & $0$     & $0$              & $0$              & $0$              & $0$              & $\ee{1.410}{2}$ \\
\bottomrule
\end{tabular}
\end{adjustbox}
\caption{Six energy-group macroscopic cross sections of the solid Hastelloy reflector (no fission, i.e. $\Sigma_{f,g}^0=\bar\nu_g=\bar E_{f,g}=0$, and no Doppler feedback). Cross sections and $1/v_g$ are in m$^{-1}$ and s\,m$^{-1}$, $D_{n,g}$ in m.}
\label{tab:params-neutronics-refl}
\end{table*}

\begin{table*}[tp]
\centering
\footnotesize
\begin{minipage}{0.55\linewidth}
\centering
\begin{tabular}{ccc}
\toprule
Group $k$ & $\beta_{d,k}$ & $\lambda_k$ [s$^{-1}$] \\
\midrule
1 & $\ee{1.224}{-4}$ & $\ee{1.247}{-2}$ \\
2 & $\ee{7.570}{-4}$ & $\ee{2.829}{-2}$ \\
3 & $\ee{3.760}{-4}$ & $\ee{4.252}{-2}$ \\
4 & $\ee{8.137}{-4}$ & $\ee{1.330}{-1}$ \\
5 & $\ee{1.479}{-3}$ & $\ee{2.925}{-1}$ \\
6 & $\ee{5.199}{-4}$ & $\ee{2.925}{-1}$ \\
7 & $\ee{4.629}{-4}$ & $1.635$ \\
8 & $\ee{1.574}{-4}$ & $3.555$ \\
\midrule
$\sum_k\beta_{d,k}$ & $\ee{4.688}{-3}$ & \\
\bottomrule
\end{tabular}
\end{minipage}%
\begin{minipage}{0.45\linewidth}
\centering
\begin{tabular}{ccc}
\toprule
Group $i$ & $\beta_{h,i}$ & $\lambda_{h,i}$ [s$^{-1}$] \\
\midrule
1 & $\ee{1.860}{-2}$ & $\ee{3.580}{-4}$ \\
2 & $\ee{1.290}{-2}$ & $\ee{1.680}{-2}$ \\
3 & $\ee{1.170}{-2}$ & $\ee{1.973}{-1}$ \\
\bottomrule
\end{tabular}
\end{minipage}
\caption{Kinetic data of the eight delayed neutron precursor groups (left, decay constant $\lambda_k$ and delayed fraction $\beta_{d,k}$) and of the three decay heat precursor groups (right, decay constant $\lambda_{h,i}$ and heat fraction $\beta_{h,i}$). The precursor fractions and decay constants are zero in the solid reflector region.}
\label{tab:params-precursors}
\end{table*}

The complete case set-up, including the mesh generation, the boundary conditions and the full-precision values of all the coefficients reported above, is stored in the following Github repository, \href{https://github.com/ERMETE-Lab/msfr-multiregion}{github.com/ERMETE-Lab/msfr-multiregion}, from which the full-order simulation can be reproduced.

\bibliographystyle{unsrt}
\bibliography{bibliography}

\end{document}

%% file: z_vardef.tex
\DeclareMathAlphabet{\mathpzc}{OT1}{pzc}{m}{it}
 

%% file: nomenclature.tex
\nomenclature[L]{$\vec{y}$}{Measurement vector}
\nomenclature[L]{$t$}{Time\nomunit{[s]}}
\nomenclature[L]{$\Delta t$}{Time Step\nomunit{[s]}}
\nomenclature[L]{$N_t$}{Number of time snapshots}
\nomenclature[L]{$N_p$}{Number of parameters}
\nomenclature[L]{$v_r$}{Reduced/Modal coefficient of rank $r$}
\nomenclature[L]{$r$}{Rank of the SVD}
\nomenclature[L]{$T$}{Temperature\nomunit{[K]}}
\nomenclature[L]{$p = \tilde{p}/\rho$}{Kinematic Pressure (Thermodynamic pressure over density)\nomunit{[m$^2$/s$^2$]}}
\nomenclature[L]{$\vec{u}$}{Velocity vector\nomunit{[m/s]}}
\nomenclature[L]{$\mathcal{V}$}{Full-Order state space}
\nomenclature[L]{$\mathcal{N}_h$}{Spatial degrees of freedom}
\nomenclature[L]{$\mathbb{X}_\psi$}{Snapshot matrix for generic field $\psi$}
\nomenclature[L]{$\hat{\mathbb{X}}_\psi$}{Reconstructed Snapshot matrix with SHRED for generic field $\psi$}
\nomenclature[L]{$\mathbb{U}_\psi$}{SVD basis for generic field $\psi$}
\nomenclature[L]{$\mathbb{V}_\psi$}{SVD reduced dynamics for generic field $\psi$}
\nomenclature[L]{$\vec{v}_j$}{Reduced state space vector at time $t_j$}
\nomenclature[L]{$\vec{x}$}{Space coordinate\nomunit{[m]}}
\nomenclature[L]{$\mathcal{N}$}{Gaussian Distribution}
\nomenclature[L]{$L$}{Number of SHRED models}
\nomenclature[L]{$c_k$}{$k$-th precursors group\nomunit{[atoms/cm$^3$]}}
\nomenclature[L]{$e$}{Test error of the SHRED output}

\nomenclature[G]{$\Omega$}{Physical Domain}
\nomenclature[G]{$\delta$}{Dirac's delta}
\nomenclature[G]{$\kappa-\varepsilon$}{Turbulent Kinetic Energy and Turbulent Dissipation Rate\nomunit{[m$^2$/s$^2$ - m$^2$/s$^3$]}}
\nomenclature[G]{$\phi_g$}{$g$-th Neutron group Flux\nomunit{[neutrons/cm$^2$/s]}}
\nomenclature[G]{$\psi$}{Generic Field}
\nomenclature[G]{$\hat{\psi}$}{SHRED reconstruction of a Generic Field}
\nomenclature[G]{$\epsilon$}{Random Noise}
\nomenclature[G]{$\boldsymbol{\mu}$}{Parameter}
\nomenclature[G]{$\tau$}{Time constant of the ULOFF scenario\nomunit{[s]}}
\nomenclature[G]{$\Phi$}{Total Neutron Flux\nomunit{[neutrons/cm$^2$/s]}}
\nomenclature[G]{$\sigma$}{Standard deviation of random gaussian noise}
\nomenclature[G]{$\varepsilon_2$}{Relative Error between FOM and SHRED in energy norm}
\nomenclature[G]{$\xi_L$}{Standard deviation of the ensembled SHRED output (L models)}

\nomenclature[A]{AI}{Artificial Intelligence}
\nomenclature[A]{ML}{Machine Learning}
\nomenclature[A]{MSFR}{Molten Salt Fast Reactor}
\nomenclature[A]{ROM}{Reduced Order Modelling}
\nomenclature[A]{PDE}{Partial Differential Equation}
\nomenclature[A]{POD}{Proper Orthogonal Decomposition}
\nomenclature[A]{SHRED}{SHallow REcurrent Decoder}
\nomenclature[A]{EVOL}{Evaluation and Viability of Liquid Fuel Fast Reactor System}
\nomenclature[A]{LSTM}{Long Short-Term Memory}
\nomenclature[A]{SDN}{Shallow Decoder Network}
\nomenclature[A]{SVD}{Singular Value Decomposition}
\nomenclature[A]{RANS}{Reynolds-Averaged Navier-Stokes}
\nomenclature[A]{ULOFF}{Unprotected Loss of Fuel Flow}
\nomenclature[A]{FOM}{Full Order Model}